\documentclass[final]{cvpr}

\usepackage{times}
\usepackage{epsfig}
\usepackage{graphicx}
\usepackage{amsmath}
\usepackage{amssymb}

\usepackage{comment}
\usepackage[utf8]{inputenc} 
\usepackage{url}            
\usepackage{booktabs}       
\usepackage{float}
\usepackage{amsfonts}       
\usepackage{nicefrac}       
\usepackage{microtype}
\usepackage{graphicx}
\usepackage{subfigure}
\usepackage{amsmath}
\usepackage{booktabs} 
\usepackage{bm}
\usepackage{amssymb}
\usepackage{soul}
\newtheorem{theorem}{Theorem}
\newtheorem{definition}{Definition}

\newtheorem{proposition}{Proposition}

\usepackage[pagebackref=true,breaklinks=true,colorlinks,bookmarks=false]{hyperref}

\def\cvprPaperID{11307} 
\def\confYear{CVPR 2021}

\begin{document}

\title{CausalVAE: Structured Causal Disentanglement in Variational Autoencoder
}

\author{Mengyue Yang$^{1,2}$, Furui Liu$^{1,}$\thanks{Corresponding author.} , Zhitang Chen$^{1}$, Xinwei Shen$^{3}$, Jianye Hao$^{1}$, Jun Wang$^{2}$\\
  $^{1}$ Noah's Ark Lab, Huawei, Shenzhen, China \\
  $^{2}$ University College London, London, United Kingdom\\
  $^{3}$ The Hong Kong University of Science and Technology, Hong Kong, China\\
  {\tt\small mengyue.yang.20@ucl.ac.uk} \\
  {\tt\small \{liufurui2,chenzhitang2,haojianye\}@huawei.com}\\
  {\tt\small xshenal@connect.ust.hk}\\ {\tt\small jun.wang@cs.ucl.ac.uk} 
}

\maketitle
\pagestyle{empty}  
\thispagestyle{empty} 

\begin{abstract}
Learning disentanglement aims at finding a low dimensional representation which consists of multiple explanatory and generative factors of the observational data. 
The framework of variational autoencoder (VAE) is commonly used to  disentangle  independent factors from observations. 
However, in real scenarios, factors with semantics are not necessarily independent. Instead, there might be an underlying causal structure which renders these factors dependent. We thus propose a new VAE based framework named CausalVAE, which includes a Causal Layer to transform independent exogenous factors into causal endogenous ones that correspond to causally related concepts in data.  We further analyze the model identifiabitily, showing that the proposed model learned from observations recovers the true one up to a certain degree by providing supervision signals (e.g. feature labels). Experiments are conducted on various datasets, including synthetic and real word benchmark CelebA.
Results show that the causal representations learned by CausalVAE are semantically interpretable, and their causal relationship as a Directed Acyclic Graph (DAG) is identified with good accuracy. Furthermore, we demonstrate that the proposed CausalVAE model is able to generate counterfactual data through ``do-operation" to the causal factors. 
\end{abstract}

\section{Introduction}

Disentangled representation learning 
is of great importance in various applications such as computer vision, speech and natural language processing, and recommender systems \cite{hsu2017unsupervised,ma2019learning,hsieh2018learning}.
The reason is that it might help enhance the performance of models, i.e. improving the generalizability, robustness against adversarial attacks as well as the explanability,  by learning data's latent disentangled representation.
One of the most common frameworks for disentangled representation learning is  Variational Autoencoders (VAE), a  deep generative model trained to  disentangle the underlying explanatory factors. Disentanglement via VAE can be achieved by a regularization term of the Kullback-Leibler (KL) divergence between the posterior of the latent factors and a standard Multivariate Gaussian prior, which enforces the learned latent factors to be as independent as possible. It is expected to recover the latent variables if the observation in real world is generated by countable independent factors. To further enhance the independence, various extensions of VAE consider minimizing the mutual information among latent factors. For example, Higgins  \etal~\cite{higgins2017beta} and Burgess  \etal~\cite{burgess2018understanding} increased the weight of the KL divergence term to enforce independence. Kim \etal~\cite{kim2018disentangling,chen2018isolating} further encourage the independence by reducing total correlation among factors.


Most existing works of disentangled representation learning make a common assumption that the real world observations are generated by countable independent factors. Nevertheless we argue that in many real world applications, latent factors with semantics of interest are causally related and thus we need a new framework that supports causal disentanglement. 

\begin{figure*}
\centering
\subfigure{
\begin{minipage}[t]{0.23\columnwidth}
\centering
\includegraphics[width=1.2\textwidth]{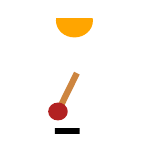}
\end{minipage}%
}
\subfigure{
\begin{minipage}[t]{0.23\columnwidth}
\centering
\includegraphics[width=1.2\textwidth]{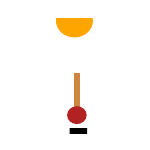}
\end{minipage}%
}
\subfigure{
\begin{minipage}[t]{0.23\columnwidth}
\centering
\includegraphics[width=1.2\textwidth]{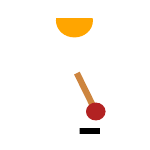}
\end{minipage}%
}
\caption{A swinging pendulum: an illustrative example}
\label{true_pendulum}
\vskip -0.2in
\end{figure*}

Consider a toy example of a swinging pendulum in Fig.~\ref{true_pendulum}. The position of the illumination source and the angle of the pendulum are causes of the position and the length of the shadow. Through causal disentangled representation learning, we aim at learning representations that correspond to the above four concepts. Obviously, these concepts are not independent and existing methods may fail to extract those factors. Furthermore, causal disentanglement allow us to manipulate the causal system to generate counterfactual data. For example, we can manipulate the latent code of shadow to create new pictures without shadow even there are pendulum and light. This corresponds to the "do-operation" \cite{pearl2009causality} in causality, where the system operates under the condition that certain variables are controlled by external forces. A deep generative model that supports "do-operation" is of tremendous value as it allows us to ask ``what-if" questions when making decisions.




In this paper, we propose a VAE-based causal disentangled representation learning framework by introducing a novel Structural Causal Model layer (Mask Layer), which allows us to recover the latent factors with semantics and structure via a causal DAG.
The input signal passes through an encoder to obtain independent exogenous factors and then a Causal Layer to generate causal representation which is taken by the decoder to reconstruct the original input. We call the whole process Causal Disentangled Representation Learning. Unlike unsupervised disentangled representation learning of which the feasibility is questionable \cite{locatello2018challenging}, additional information is required as weak supervision signals to achieve causal representation learning.  By ``weak supervision", we emphasize that in our work, the causal structure of the latent factors is automatically learned, instead of being given as a prior in \cite{causalgan}. To train our model, we propose a new loss function which includes the VAE evidence lower bound loss and an acyclicity constraint imposed on the learned causal graph to guarantee its ``DAGness". In addition, we analyze the identifiablilty of the proposed model, showing that the learned parameters of the disentangled model recover the true one up to certain degree.  The contribution of our paper is three-fold.
(1) We propose a new framework named CausalVAE that supports causal disentanglement and ``do-operation"; (2) Theoretical justification on model identifiability is provided;  (3) We conduct comprehensive experiments with synthetic and real world face images to demonstrate that the learned factors are with causal semantics and can be intervened to generate counterfactual images that do not appear in training data.
 
    

\section{Related Works} 
In this section, we review state-of-the-art disentangled representation learning methods, including some recent advances on combining causality and disentangled representation learning. We also present preliminaries of causal structure learning from pure observations which is a key ingredient of our proposed CausalVAE framework.


\vspace{-2mm}

\subsection{Disentangled Representation Learning}
Conventional disentangled representation learning methods learn mutually independent latent factors by an encoder-decoder framework. In this process, a standard normal distribution is used  as a prior of the latent code.  
A variational posterior $q(\mathbf{z}|\mathbf{x})$ is then used to approximate the unknown true posterior $p(\mathbf{z}|\mathbf{x})$.
This framework was further extended by adding new independence regularization terms to the original loss function, leading to various algorithms. $\beta$-VAE \cite{higgins2017beta} proposes an adaptation framework which adjusts the weight of KL term to balance between  independence of disentangled factors and the reconstruction performance. While factor VAE \cite{chen2018isolating} proposes a new  framework which focuses solely on the independence of factors. Ladder VAE \cite{laddervae} on the other hand, leverages the structure of ladder neural network  to train a structured VAE for hierarchical disentanglement.
\begin{figure*}[h]
\begin{center}
\centerline{\includegraphics[width=2\columnwidth]{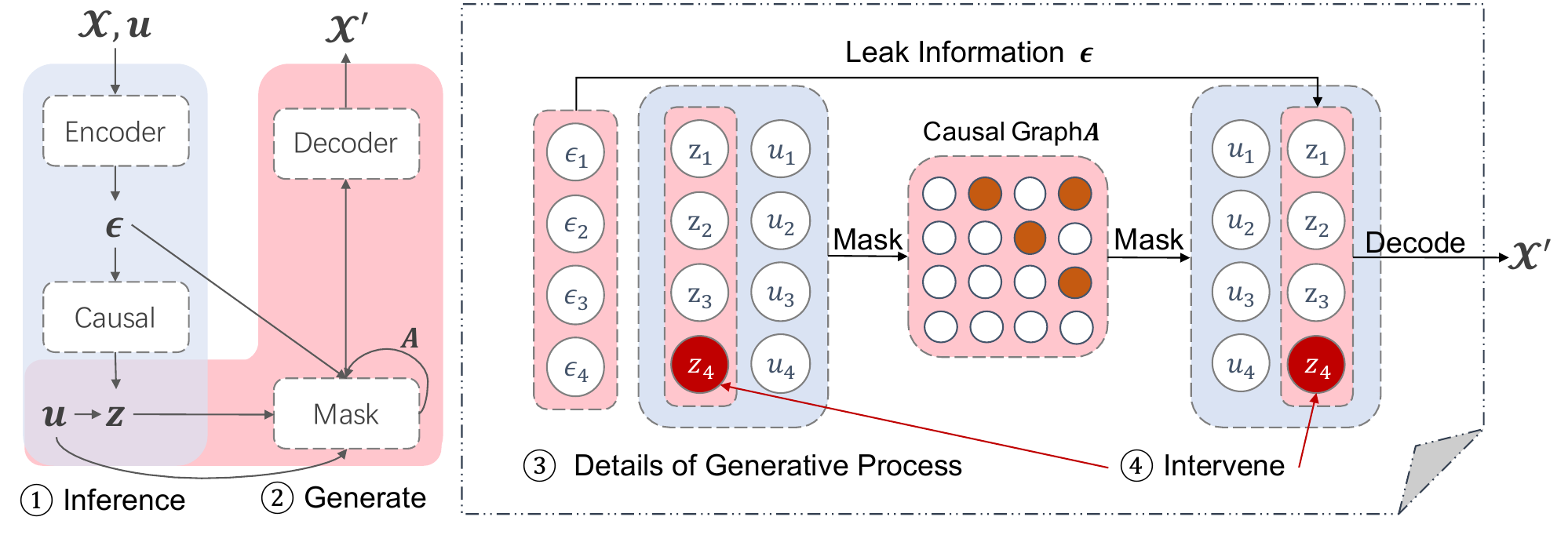}}
\caption{Model structure of CausalVAE. The encoder takes observation $\mathbf{x}$ as inputs to generate independent exogenous variable $\bm{\epsilon}$, whose prior distribution is assumed to be standard Multivariate Gaussian. Then it is transformed by the Causal Layer into causal representations $\mathbf{z}$ (Eq. \ref{sem}) with a conditional prior distribution $p(\mathbf{z}|\mathbf{u})$.  A Mask Layer is then applied to $\mathbf{z}$ to resemble the SCM in Eq. \ref{maskedlayer}. After that, $\mathbf{z}$ is taken as the input of the decoder to 
reconstruct the observation $\mathbf{x}$. }
\label{cvae_structure}
\end{center}

\end{figure*}
Nevertheless the aforementioned unsupervised disentangled representation learning algorithms do not perform well in some situations where there is complex causal relationship among factors. Furthermore, they are challenged for lacking inductive bias and thus the model identifiability cannot be guaranteed \cite{locatello2018challenging}. The identifiability problem of VAE is defined as follows: 
if the parameters $\tilde{\bm{\theta}}$ learned from data lead to a marginal distribution equal to the true one parameterized by $\bm{\theta}$, i.e., $p_{\tilde{\bm{\theta}}}(\mathbf{x}) = p_{\bm{\theta}}(\mathbf{x}) $, then the joint distributions also match, i.e.  $p_{\tilde{\bm{\theta}}}(\mathbf{x},\mathbf{z}) = p_{\bm{\theta}}(\mathbf{x},\mathbf{z}) $. Therefore, the 
rotation invariance of prior $p(\mathbf{z})$ (standard Multivariate Gaussian distribution) will lead the unindentifiable of $p(\mathbf{z})$. Khemakhem   \etal~\cite{nonlinearica} prove that there is infinite number of distinct models entailing the same joint distributions, which means that the underlying generative model is not identifiable through unsupervised learning. On the contrary, by leveraging a few labels, one is able to recover the true model \cite{mathieu2018disentangling,locatello2018challenging}. Kulkarni \etal~\cite{kulkarni2015deep} and Locatello \etal~\cite{locatello2019disentangling} use additional labels to reduce the model ambiguity. Khemakhem \etal~\cite{nonlinearica} gives an identifiability of VAE with additional inputs, by leveraging the theory of nonlinear Independent Component Analysis (nonlinear ICA) \cite{brakel2017learning}. 

\subsection{Causal Discovery \& Causal Disentangled Representation Learning}

We refer to causal representation as ones structured by a causal graph. Discovering the causal graph from pure observations has attracted  large amounts of attention in the past decades \cite{hoyer2009nonlinear,zhang2012identifiability,shimizu2006linear}. Methods for causal discovery use either observational data or a combination of observational and interventional data. We first introduce a set of methods based on observational data. Pearl \etal~\cite{pearl2009causality} introduced a Probabilistic Graphical Models (PGMs) based language to describe causality among variables. Shimizu \etal~\cite{shimizu2006linear} proposed an effective method called LiNGAM to learn the causal graph and they prove the model identifiability under the linearity and non-Gaussianity assumption. Zheng  \etal~\cite{zheng2018dags} proposed NOTEARs with a fully differentiable DAG constraint for causal structure learning, which drastically reduces a very complicated combinatorial optimization problem to a continuous optimization problem. Zhu \etal~\cite{zhu2019causal} proposed a flexible and efficient Reinforcement Learning (RL) based method to search over a DAG space for a best graph with a highest score. When interventions are doable, that is, one can manipulate the causal system and collect data under interventions, methods are proposed for causal discovery.  Tillman  \etal~\cite{DBLP:journals/jmlr/TillmanS11, DBLP:journals/corr/abs-1302-6815} show the identifiability of learned causal structure from interventional data. Peters \etal~\cite{DBLP:journals/corr/abs-1910-01075, peters2015causal,peters2017elements} explores the structure invariance across multiple domains under interventions to identify causal edges.

Recently, the community has raised interest of combining causality and disentangled representation.  Suter  \etal~\cite{suter2018robustly} used causality to explain disentangled latent representations. Kocaoglu \etal~\cite{causalgan} proposed a method called CausalGAN which supports "do-operation" on images but it requires the causal graph given as a prior. Instead of assuming independent latent factors, Besserve \etal~\cite{besserve2018counterfactuals} adopts dependent latent factors in the model. It relies on the principle of ``independence mechanism" or modularity for disentanglement, and design a layer containing a few non-structured nodes, representing outputs of mutually independent causal mechanisms \cite{peters2017elements},  which contribute together to the final predictions to achieve disentanglement. In our model, we disentangle factors by causally structured layers (masking layer), and the model structure is different from theirs.   Sch{\"o}lkopf \etal~\cite{scholkopf2019causality} claims the importance and necessity of causal disentangled representation learning but it still remains conceptual. To the best of our knowledge, our work is the first one that successfully implements the idea of causal disentanglement.
\vspace{-3mm} 

\section{Causal Disentanglement in Variational Autoencoder}

We start with the definition of causal representation, and then propose a new framework to achieve causal disentanglement by leveraging additional inputs, e.g. labels of concepts. Firstly, we give an overview of our proposed CausalVAE model structure in Fig. \ref{cvae_structure}. A Causal Layer, which essentially describes a Structural Causal Model (SCM) \cite{shimizu2006linear}, is introduced to a conventional VAE network. The Causal Layer transforms the independent exogenous factors to causal endogenous factors corresponding to causally related concepts of interest. A mask mechanism \cite{ng2019masked} is then used to propagate the effect of parental variables to their children, mimicking the assignment operation of SCMs. Such a Causal Layer is the key to supporting intervention or ``do-operation" to the system. 









\subsection{Transforming Independent Exogenous Factors into Causal Representations}

 Our model is within the framework of VAE-based disentanglement. In addition to the encoder and the decoder structures, we introduce a Structural Causal Model (SCM) layer to learn causal representations. To formalize causal representation, we consider $n$ concepts of interest in data.  The concepts in observations are causally structured by a Directed Acyclic Graph (DAG) with an adjacency matrix $\mathbf{A}$. Though a general nonlinear SCM is preferred, for simplicity, in this work, the Causal Layer exactly implements a Linear SCM as described in Eq. \ref{sem} (shown in Fig. \ref{cvae_structure} \textcircled{1}), 
 \begin{equation}
    \mathbf{z} = \mathbf{A}^T\mathbf{z}+\bm{\epsilon} = (I-\mathbf{A}^T)^{-1}\bm{\epsilon},\\
    \bm{\epsilon}\sim \mathcal{N}(\mathbf{0},\mathbf{I}),
    \label{sem}
\end{equation}
where $\mathbf{A}$ is the parameters to be learnt in this layer. $\bm{\epsilon}$ are independent Gaussian exogenous factors and $\textbf{z}\in \mathbb{R}^{n}$ is structured causal representation of $n$ concepts that is generated by a DAG and thus $\mathbf{A}$ can be permuted into a strictly upper triangular matrix.
 

Unsupervised learning of the model might be infeasible due to the identifiability issue as discussed in \cite{locatello2018challenging}. To address this problem, similar to iVAE \cite{nonlinearica}, we adopt additional information $\mathbf{u}$ associated with the true causal concepts as supervising signals. In our work, we use the labels of the concepts. The additional information $\mathbf{u}$ is utilized in two ways. Firstly, we propose a conditional prior $p(\mathbf{z|u})$ to regularize the learned posterior of $\mathbf{z}$. This guarantees that the learned model belongs to an identifiable family. Secondly, we also leverage $\mathbf{u}$ to learn the causal structure $\mathbf{A}$. Besides learning the causal representations, we further enable the model to support intervention to the causal system to generate counterfactual data which does not exist in the training data. 



\subsection{Structural Causal Model Layer}\label{mask_layer}

Once the causal representation $\textbf{z}$ is obtained, it passes through a Mask Layer \cite{ng2019masked} to reconstruct itself. Note that this step resembles a SCM which depicts how children are generated by their corresponding parental variables. We will show why such a layer is necessary to achieve intervention. Let $z_i$ be the $i$th variable in the vector $\mathbf{z}$. The adjacency matrix associated with the causal graph is $\mathbf{A} = [\mathbf{A}_1|\dots|\mathbf{A}_n]$ where $\mathbf{A}_i \in \mathbb{R}^n$ is the weight vector such that $A_{ji}$ encodes the causal strength from $z_j$ to $z_i$.  We have a set of mild nonlinear and invertible functions $ [g_1, g_2,\dots,g_n]$ that map parental variables to the child variable. Then we write
\begin{align}\label{maskedlayer}
    z_i = g_i(\mathbf{A}_i \circ \mathbf{z}; \bm{\eta}_i) + \epsilon_i,
\end{align}
where $\circ$ is the element-wise multiplication and $\bm{\eta}_i$ is the parameter of $g_i(\cdot)$  (as shown in Fig. \ref{cvae_structure} \textcircled{3}).  Note that according to Eq. \ref{sem}, we can simply write $z_i = \mathbf{A}_i^T \mathbf{z}+ \epsilon_i$. However, we find that adding a mild nonlinear function $g_i$ results in more stable performances. To show how this masking works, consider a variable $z_i$ and $\mathbf{A}_i \circ \mathbf{z}$  equals a vector that only contains its parental information as it masks out all $z_i$'s non-parent variables. By minimizing the reconstruction error, the adjacency matrix $\mathbf{A}$ and the parameter $\bm{\eta}_i$ of the mild nonlinear function $g_i$ are trained. 

This layer makes intervention or "do-operation" possible. Intervention \cite{pearl2009causality} in causality refers to modifying a certain part of a system by external forces and one is interested in the outcome of such manipulation. To intervene $z_i$, we set $z_i$ on the RHS of Eq. \ref{maskedlayer} (corresponding to the $i-$th node of $\mathbf{z}$ in the first layer in Fig. \ref{cvae_structure}) to a fixed value, and then its effect is delivered to all its children as well as itself on the LHS of Eq. \ref{maskedlayer} (corresponding to some nodes of $\mathbf{z}$ in the second layer). 
Note that intervening the cause will change the effect, whereas intervening the effect, on the other hand, does not change the cause because information can only flow into the next layer from the previous one in our model, which is aligned with the definition of causal effects.

\subsection{A Probabilistic Generative Model for CausalVAE}\label{prob_analysis}

We give a probabilistic formulation of the proposed generative model (shown in Fig. \ref{cvae_structure} \textcircled{2}). 
Denote by $\mathbf{x}\in \mathbb{R}^{d}$ the observed variables and $\mathbf{u}\in \mathbb{R}^{n}$ the additional information. $u_i$ is the label of the $i$-th concept of interest in data. Let $\bm{\epsilon} \in \mathbb{R}^n$  be the latent exogenous independent variables and $\mathbf{z}\in \mathbb{R}^{n}$ be the latent endogenous variables with semantics where $\mathbf{z} = \mathbf{A}^{T}\mathbf{z} + \bm{\epsilon} = (\mathbf{I} - \mathbf{A}^T)^{-1}\bm{\epsilon}$. For simplicity, we denote $\mathbf{C}=(\mathbf{I} - \mathbf{A}^T)^{-1}$. 

We treat both $\mathbf{z}$ and $\bm{\epsilon}$ as latent variables. Consider the following conditional generative model parameterized by $\bm{\theta} = (\mathbf{f}, \mathbf{h},\mathbf{C}, \mathbf{T}, \bm{\lambda})$:
\begin{align}
    p_{\bm{\theta}}(\mathbf{x,z,\bm{\epsilon}|u}) &= p_{\bm{\theta}}(\mathbf{x|z,\bm{\epsilon},u})p_{\bm{\theta}}(\mathbf{\bm{\epsilon},z|u}). \label{generator}
\end{align}


Let $\mathbf{f}(\mathbf{z})$ denote the decoder which is assumed to be an invertible function and $\mathbf{h}(\mathbf{x}, \mathbf{u})$ denotes the encoder. We define the generative and inference models as follows:
\begin{equation}\label{Eq: conditional}
\begin{split}
    p_{\bm{\theta}}(\mathbf{x|z,\bm{\epsilon},u})=p_{\bm{\theta}}(\mathbf{x}|\mathbf{z}) \equiv p_{\bm{\xi}}(\mathbf{x} - \mathbf{f}(\mathbf{z})),\\ q_{\bm{\phi}}(\mathbf{z},\bm{\epsilon}|\mathbf{x},\mathbf{u})\equiv q(\mathbf{z}|\bm{\epsilon})q_{\bm{\zeta}}(\bm{\epsilon} - \mathbf{h}(\mathbf{x},\mathbf{u})),
\end{split}
\end{equation}
which is obtained by assuming the following decoding and encoding processes:
\begin{equation}\label{xi}
     \mathbf{x}=\mathbf{f}(\mathbf{z})+\bm{\xi},~~~\bm{\epsilon}=\mathbf{h(x,u)}+\bm{\zeta},
\end{equation}
where $\bm{\xi}$ and $\bm{\zeta}$ are the vectors of independent noise with probability densities $p_{\bm{\xi}}$ and $q_{\bm{\zeta}}$. When $\bm{\xi}$ and $\bm{\zeta}$ are infinitesimal, the encoder and decoder can be regarded as deterministic ones. We define the joint prior $p_{\bm{\theta}}(\bm{\epsilon},\mathbf{z}|\mathbf{u})$ for latent variables $\mathbf{z}$ and $\bm{\epsilon}$ as
\begin{equation}\label{eq:joint_prior}
    p_{\bm{\theta}}(\bm{\epsilon},\mathbf{z}|\mathbf{u}) = p_{\bm{\epsilon}}(\bm{\epsilon})p_{\bm{\theta}}(\mathbf{z|u}),
\end{equation}
where $p_{\bm{\epsilon}}(\bm{\epsilon})=\mathcal{N}(\mathbf{0},\mathbf{I})$ and the prior of latent endogenous variables $p_{\bm{\theta}}(\mathbf{z|u})$ is a factorized Gaussian distribution conditioning on the additional observation $\mathbf{u}$, i.e.
\begin{align}
    p_{\bm{\theta}}(\mathbf{z}|\mathbf{u}) = \Pi_i^n p_{\bm{\theta}}(z_i|u_i), p_{\bm{\theta}}(z_i|u_i) = \mathcal{N}(\lambda_1(u_i),\lambda_2^2(u_i)), \label{prior}
\end{align}
where $\lambda_1$ and $\lambda_2$ are an arbitrary functions. In this paper, we let $\lambda_1(\mathbf{u}) = \mathbf{u}$ and $\lambda_2(\bm{u})\equiv 1$. 
The distribution has two sufficient statistics, the mean and variance of $\mathbf{z}$, which are denoted by sufficient statistics $\mathbf{T}(\mathbf{z}) = (\bm{\mu}(\mathbf{z}), \bm{\sigma}(\mathbf{z}))=(T_{1,1}(z_1),\dots,T_{n,2}(z_n))$. We use these notations for model idnetifiability analysis in Section \ref{identifiability_analysis}. 








\section{Learning Strategy}\label{learning_strategy}

In this section, we discuss how to train the CausalVAE model in order to learn the causal representation as well as the causal graph simultaneously.
\subsection{Evidence Lower Bound of CausalVAE}
We apply variational Bayes to learn a tractable distribution $q_\phi(\bm{\epsilon},\mathbf{z}|\mathbf{x},\mathbf{u})$ to approximate the true posterior $p_{\bm{\theta}}(\bm{\epsilon},\mathbf{z}|\mathbf{x},\mathbf{u})$. 
Given data set $\mathcal{X}$ with the empirical data distribution $q_\mathcal{X}(\mathbf{x},\mathbf{u})$, the parameters $\bm{\theta}$ and $\bm{\phi}$ are learned by optimizing the following evidence lower bound (ELBO):
\begin{align}
     \mathbb{E}_{q_\mathcal{X}}[\log p_{\bm{\theta}}(\mathbf{x}|\mathbf{u})] \ge \text{ELBO}  
    &= \mathbb{E}_{q_\mathcal{X}}[\mathbb{E}_{\bm{\epsilon},\mathbf{z}\sim q_{\bm{\phi}}}[\log p_{\bm{\theta}}(\mathbf{x}|\mathbf{z},\bm{\epsilon},\mathbf{u})]\nonumber\\&-\mathcal{D}(q_{\bm{\phi}} (\bm{\epsilon},\mathbf{z}|\mathbf{x},\mathbf{u})||p_{\bm{\theta}}(\bm{\epsilon},\mathbf{z}|\mathbf{u}))], \label{elbo}
\end{align}
where $\mathcal{D}(\cdot\|\cdot)$ denotes KL divergence. Eq. \ref{elbo} is intractable in general. However, thanks to the one-to-one correspondence between $\bm{\epsilon}$ and $\mathbf{z}$, we simplify the variational posterior as follows:
\begin{align}
     \small q_{\bm{\phi}} (\bm{\epsilon},\mathbf{z}|\mathbf{x},\mathbf{u}) &= q_{\bm{\phi}} (\bm{\epsilon}|\mathbf{x},\mathbf{u})\delta(\bm{z}=\mathbf{C}\bm{\epsilon})\nonumber\\
     &=q_{\bm{\phi}} (\mathbf{z}|\mathbf{x},\mathbf{u})\delta(\bm{\epsilon}=\mathbf{C}^{-1}\bm{z}), \label{inferencemodel}
\end{align}
where $\delta(\cdot)$ is the Dirac delta function. According to the model assumptions introduced in Section \ref{prob_analysis}, i.e., generation process (Eq. \ref{Eq: conditional}) and prior (Eq. \ref{eq:joint_prior}), we attain a neat form of ELBO loss as follows:

\begin{proposition} ELBO defined in Eq. \ref{elbo} can be written as:
\begin{align}\label{eq:elbo_decom}
     \small \text{ELBO} =& \mathbb{E}_{q_\mathcal{X}}[\mathbb{E}_{q_\phi(\mathbf{z}|\mathbf{x},\mathbf{u})}[\log p_{\bm{\theta}}(\mathbf{x}|\mathbf{z})] \nonumber\\&- \mathcal{D}(q_\phi(\bm{\epsilon}|\mathbf{x},\mathbf{u})||p_{\bm{\epsilon}}(\bm{\epsilon}))\nonumber\\&-\mathcal{D}(q_\phi(\mathbf{z}|\mathbf{x},\mathbf{u}) ||  p_{\bm{\theta}}(\mathbf{z}|\mathbf{u}))].
\end{align}
\end{proposition}
Details of the proof are given in the Appendix A. With this form, we can easily implement a loss function to train the CausalVAE model. 

\subsection{Learning the Causal Structure of Latent Codes}


In addition to the encoder and decoder, our CausalVAE model involves a Causal Layer with a DAG structure to be learned. Note that both $\mathbf{z}$ and $\mathbf{A}$ are unknown, to ease the training task and guarantee the identifiability of causal graph $\mathbf{A}$, we leverage the additional labels $\mathbf{u}$ to construct the following constraint:
\begin{equation}
     \small l_u=\mathbb{E}_{q_{\mathcal{X}}}\Vert \mathbf{u} - \sigma(\mathbf{A}^T \mathbf{u})\Vert_2^2\leq \kappa_1,
    \label{constraint_A}
\end{equation}
where $\sigma$ is a logistic function as our labels are binary and $\kappa_1$ is the small positive constant value. This follows the idea that $\mathbf{A}$ should also describe the causal relations among labels well. Similarly we apply the same constraint to the learned latent code $\mathbf{z}$ as follows:
\begin{equation}\label{constraint_m}
   l_m= \mathbb{E}_{\mathbf{z}\sim q_{\bm{\phi}}}\sum_{i=1}^n \Vert z_i - g_i(\mathbf{A}_i  \circ\mathbf{z}; \bm{\eta}_i) \Vert^2 \leq \kappa_2,
\end{equation}
where $\kappa_2$ is the small positive constant value.
Lastly, the causal adjacency matrix $\mathbf{A}$ is constrained to be a DAG. Instead of using traditional DAG constraint that is combinatorial, we adopt a continuous  differentiable constraint function \cite{zheng2018dags, DBLP:journals/corr/abs-1906-04477,DBLP:journals/corr/abs-1911-07420,yu2019dag} . The function attains 0 if and only if the adjacency matrix $\mathbf{A}$ corresponds to a DAG \cite{yu2019dag}, i.e.
\begin{align}
    H(\mathbf{A}) \equiv	 tr((\mathbf{I}+\frac{c}{m}\mathbf{A}\circ \mathbf{A})^n) - n = 0, \label{dag}
\end{align}
where $c$ is an arbitrary positive number. The training procedure of our CausalVAE model reduces to the following constrained optimization:
\begin{align}
   \text{maximize}  ~~~\text{ELBO}, \nonumber \\ \text{s.t.}~~~(\ref{constraint_A})(\ref{constraint_m})(\ref{dag}).
\nonumber
\end{align}
By lagrangian multiplier method, we have the new loss function
\begin{align}
    \mathcal{L} =  -\text{ELBO}
    + \alpha H(\mathbf{A})+ \beta l_u + \gamma l_m, \label{causal_elbo} 
\end{align}
where $\alpha,\beta, \gamma$ denote regularization hyperparameters. 
\section{Identifiability Analysis}\label{identifiability_analysis}
In this section, we present the identifiability of our proposed model. We adopt the $\sim$-\textit{identifiable} \cite{nonlinearica} as follows:

\begin{definition} Let $\sim$ be the binary relation on $\Theta$ defined as follows:
\begin{equation}
\begin{split}
    (\mathbf{f},\mathbf{h},\mathbf{C}, \mathbf{T},\bm{\lambda})\sim (\tilde{\mathbf{f}},\tilde{\mathbf{h}}, \tilde{\mathbf{C}}, \tilde{\mathbf{T}},\tilde{\bm{\lambda}}) \\\Leftrightarrow \exists \mathbf{B_1},\mathbf{B_2},\mathbf{b_1},\mathbf{b_2} \vert\\ 
     \mathbf{T}(\mathbf{h}(\mathbf{x},\mathbf{u})) = \mathbf{B_1}\tilde{\mathbf{T}}(\tilde{\mathbf{h}}(\mathbf{x},\mathbf{u}))+\mathbf{b_1},\mathbf{T}(\mathbf{f^{-1}}(\mathbf{x})) \\= \mathbf{B_2}\tilde{\mathbf{T}}( {\mathbf{ \tilde{f}^{-1}}}(\mathbf{x}))+\mathbf{b_2},\forall \mathbf{x}\in \mathcal{X},
    \end{split}
\end{equation}
where $\mathbf{C}=(\mathbf{I} - \mathbf{A}^T)^{-1}$. If $\mathbf{B_1}$ is an invertible matrix and $\mathbf{B_2}$ is an invertible diagonal matrix with diagonal elements associated to $u_i$. We say that the model parameter is $\sim$-\textit{identifiable}.
\end{definition}
Following \cite{nonlinearica}, we obtain the identifiability of our causal generative model as follows.

\begin{theorem} Assume that the data we observed are generated according Eq. \ref{generator}-\ref{Eq: conditional} and the following assumptions hold, 
\begin{enumerate}
    \item The set $\{\mathbf{}x\in\mathcal{X}|\phi_\xi(\mathbf{x})=0\}$ has measure zero, where $\phi_\xi$ is the characteristic function of the density $p_\xi$ defined in Eq.  \ref{xi}.
    \item The decoder function $\mathbf{f}$ is differentiable and the Jacobian matrix of  $\mathbf{f}$ is of full rank \footnote{(rank equals to its smaller dimension)}.
    \item The sufficient statistics $T_{i,s}({z}_i)\not=0$ almost everywhere for all $1\le i \le n$ and $1\le s \le2$, where $T_{i,s}({z}_i)$ is the $s$th statistic of variable $z_i$.
    \item The additional observations $u_i\not=0$.
\end{enumerate}
 Then the parameters $(\mathbf{f}, \mathbf{h}, \mathbf{C}, \mathbf{T}, \bm{\lambda})$ are $\sim$-\textit{identifiable}. 
\end{theorem}

Although the parameters $\bm{\theta}$ of true generative model are unknown during the learning process, the identifiablity of generative model  given by Theorem 1  guarantees the parameters $\widetilde{\bm{\theta}}$ learned by hypothetical functions are in an identifiable family. This shows that the learned parameters of the generative model recover the true one up to certain degree.

In addition, all $z_i$ in $\mathbf{z}$  align to the additional observation of concept $i$ and they are expected to inherent the causal relationship of causal system. That is why that it could guarantee that the $\mathbf{z}$ are causal representation.

The identifiability of the model under supervision of additional information is obtained thanks to the conditional prior $p_{\bm{\theta}}(\mathbf{z}|\mathbf{u})$. The conditional prior guarantees that sufficient statistics of $p_{\bm{\theta}}(\mathbf{z}|\mathbf{u})$ are related to the value of $\mathbf{u}$. A complete proof of \textbf{Theorem 1} is available in Appendix B.

\section{Experiments}
In this section, we conduct experiments using both synthetic dataset and real human face image dataset and we compare our CausalVAE model against existing state of the art methods on disentangled representation learning. We focus on examing whether a certain algorithm is able to learn interpretable representations and whether outcomes of intervention on learned latent code is consistent to our understanding of the causal system. 




\begin{figure*}[h]

\begin{center}
\centerline{\includegraphics[width=2\columnwidth]{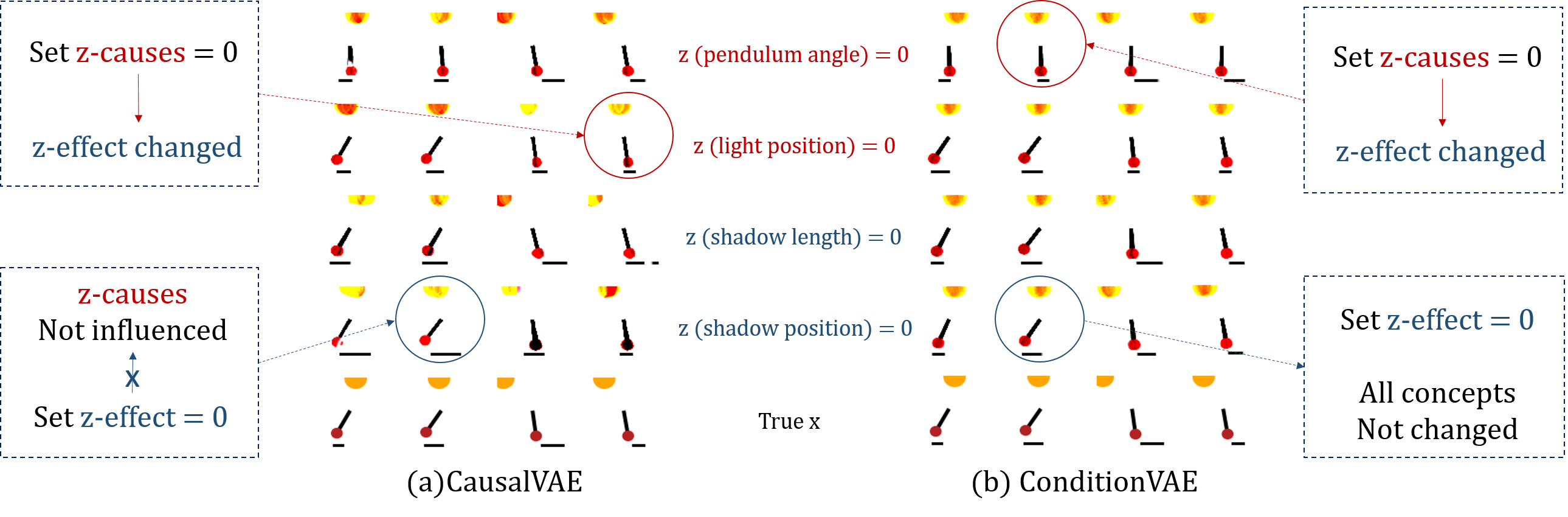}}

\caption{The results of Intervention experiments on the pendulum dataset. Each row shows the result of controlling the \textsc{pendulum angle}, \textsc{light angle}, \textsc{shadow length}, and  \textsc{shadow location} respectively. The bottom row is the original input image. More intervention results on other synthetic dataset are shown in Appendix D.3.}
\label{pendulum_res}
\end{center}

\end{figure*}


\begin{figure*}[h]
\vskip -0.3in
\begin{center}
\centerline{\includegraphics[width=1.8\columnwidth]{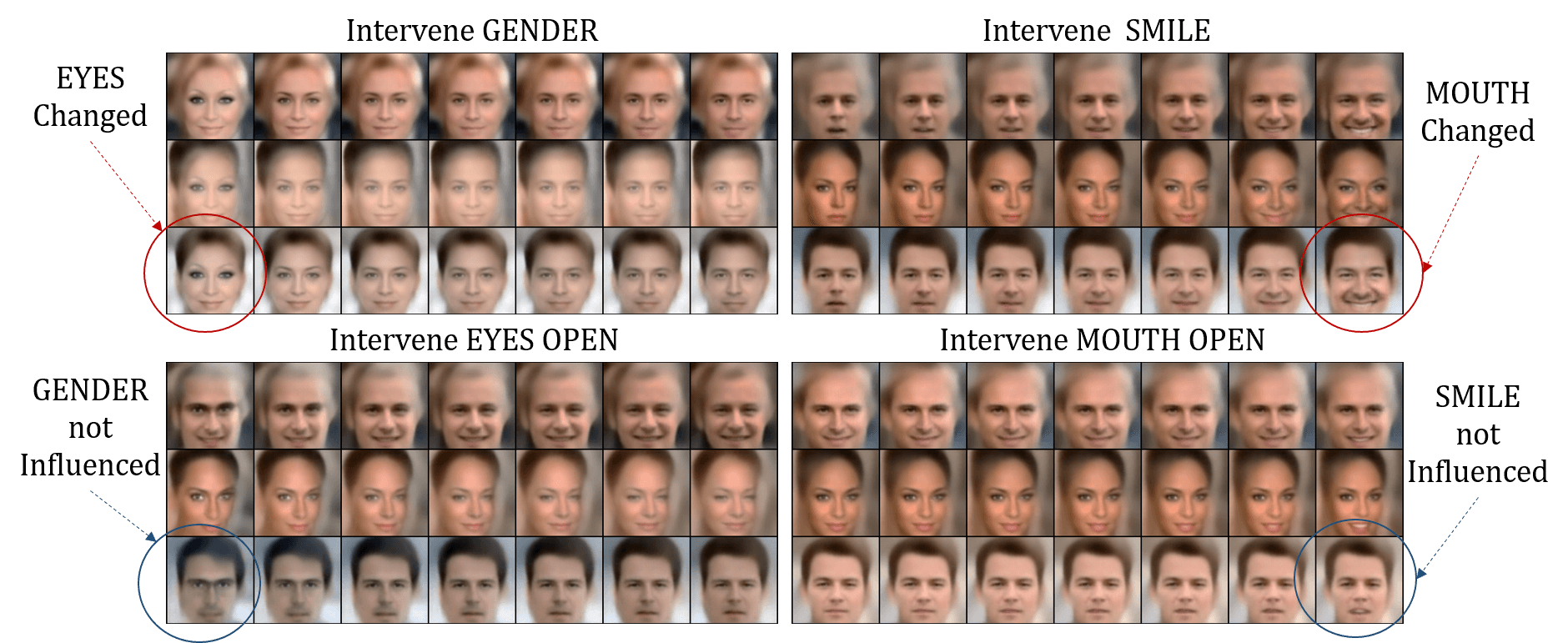}}
\vskip -0in
\caption{Results of CausalVAE model on CelebA(\textsc{Smile}). The controlled factors are \textsc{gender},\textsc{ smile, eyes open} and \textsc{mouth open} respectively. More intervention results are shown in Appendix D.3.}
\label{do_smile} 
\end{center}
\end{figure*}

\begin{figure*}[htb]
\centering
\subfigure[Initialize]{
\begin{minipage}{0.45\columnwidth}
\centering
\includegraphics[width=1\textwidth]{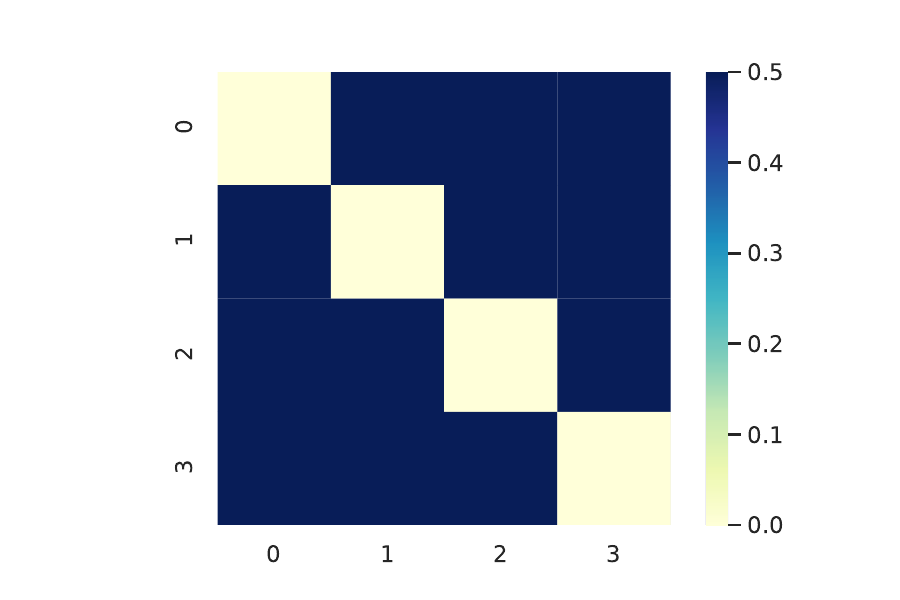}
\end{minipage}%
}
\subfigure[After 1 epoch]{
\begin{minipage}{0.45\columnwidth}
\centering
\includegraphics[width=1\textwidth]{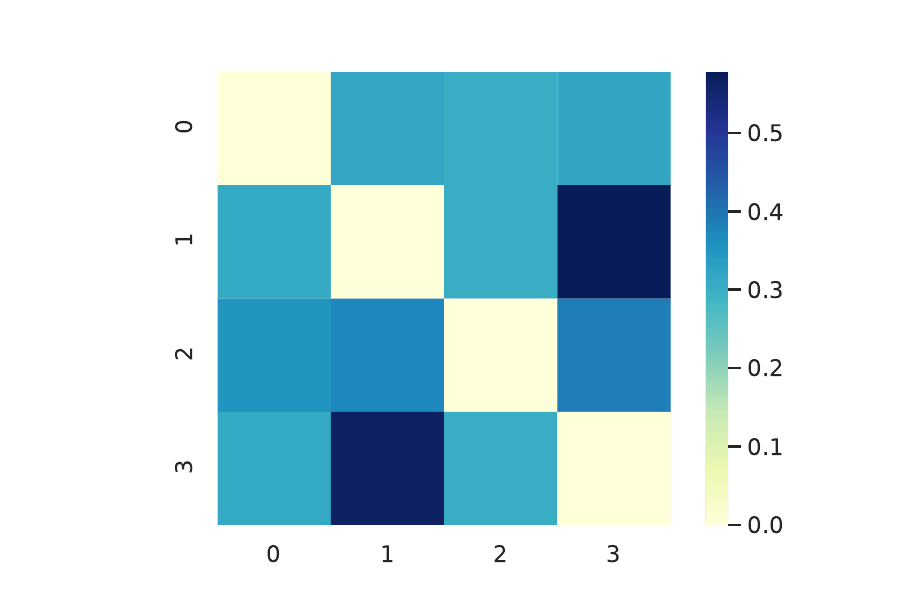}
\end{minipage}%
}
\subfigure[After 5 epochs]{
\begin{minipage}{0.45\columnwidth}
\centering
\includegraphics[width=1\textwidth]{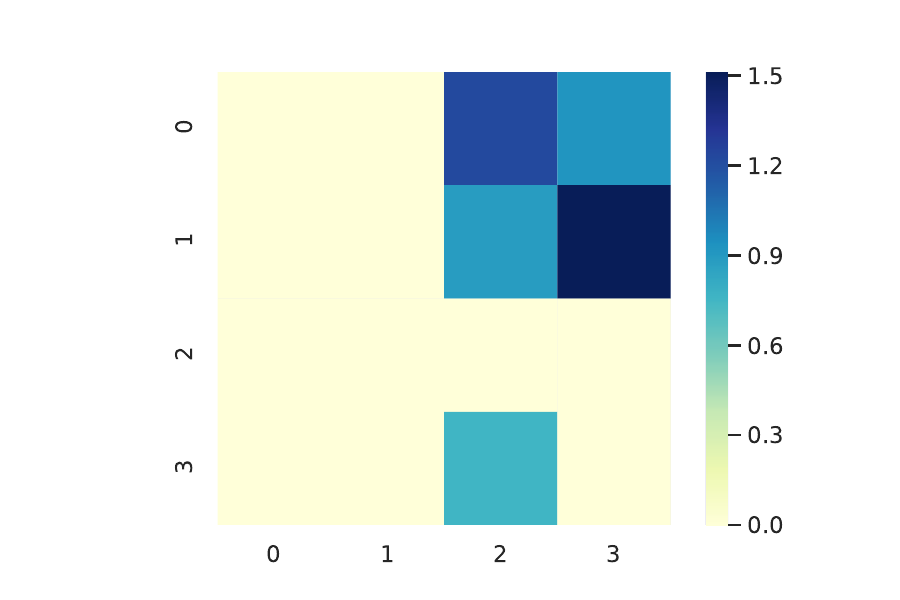}
\end{minipage}
}
\subfigure[The true]{
\begin{minipage}{0.45\columnwidth}
\centering
\includegraphics[width=1\textwidth]{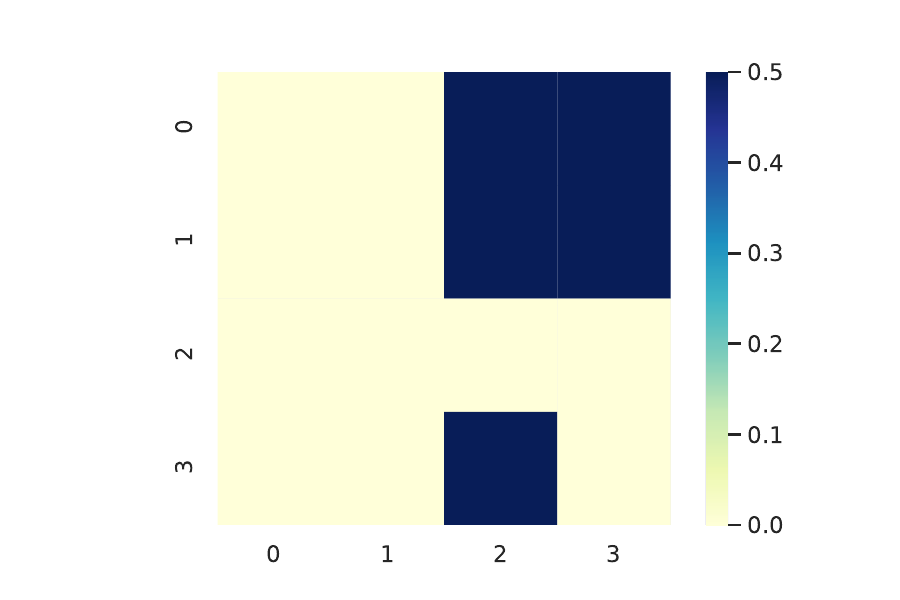}
\end{minipage}%
}
\caption{The learning process of causal matrix $\mathbf{A}$. The concepts include: \textsc{Gender}, \textsc{Smile}, \textsc{Eyes Open}, \textsc{Mouth Open} (top-to-bottom and left-to-right order); (c) converged $\mathbf{A}$, (d) ground truth .}
\label{learn_A}
\end{figure*}

\begin{table*}[h]
\centering
    \caption{The MIC and TIC between learned representation $\mathbf{z}$ and the label $\mathbf{u}$. The results show that among all compared methods, the learned factors of our proposed CausalVAE achieve best alignment to the concepts of interest. (Note: the metrics include mean $\pm$ standard errors in table.)}  \label{tab:freq1}

  \resizebox{\textwidth}{!}{
  \begin{tabular}{ccccccccccc}
    \toprule
    & \multicolumn{2}{c}{CausalVAE } & \multicolumn{2}{c}{ConditionVAE}& \multicolumn{2}{c}{$\beta$-VAE}& \multicolumn{2}{c}{CausalVAE-unsup}& \multicolumn{2}{c}{LadderVAE}\\
    \cmidrule(r){2-11}
    Metrics(\%) &MIC&TIC&MIC&TIC&MIC&TIC&MIC&TIC&MIC&TIC\\
    \cmidrule(r){1-11}
    Pendulum &\textbf{95.1 {{$\pm$2.4}}}&\textbf{81.6 {{$\pm$1.9}}} &93.8 {{$\pm$3.3}}& 80.5 {{$\pm$1.4}}&22.6 {{$\pm$4.6}} & 12.5 {{$\pm$2.2}} & 21.2 {{$\pm$1.4}} & 12.0 {{$\pm$1.0}}& 22.4 {{$\pm$3.1}} & 12.8 {{$\pm$1.2}}\\
    \cmidrule(r){1-11}
    Flow &72.1 {{$\pm$1.3}}&56.4 {{$\pm$1.6}} &\textbf{75.5 {{$\pm$2.3}}}& \textbf{56.5 {{$\pm$1.8}}}&23.6 {{$\pm$3.2}} & 12.5 {{$\pm$0.6}} & 22.8 {{$\pm$2.7}} & 12.4 {{$\pm$1.4}}& 34.3 {{$\pm$4.3}} & 24.4 {{$\pm$1.5}}\\
    \cmidrule(r){1-11}
    CelebA(\textsc{Smile})  &\textbf{83.7 {{$\pm$6.2}}}&\textbf{71.6 {{$\pm$7.2}}} &78.8 {{$\pm$10.9}}& 66.1 {{$\pm$12.1}}&22.5 {{$\pm$1.2}} & 9.92 {{$\pm$1.2}} & 27.2 {{$\pm$5.3}} & 14.6 {{$\pm$4.2}}& 23.5 {{$\pm$3.0}} & 10.3 {{$\pm$1.6}}\\
    \cmidrule(r){1-11}
    CelebA(\textsc{Beard})  &\textbf{92.3 {{$\pm$5.6}}}&\textbf{83.3 {{$\pm$8.6}}} &89.8 {{$\pm$6.2}}& 78.7 {{$\pm$7.7}}&22.4 {{$\pm$1.9}} & 9.82{{$\pm$2.2}} & 11.4 {{$\pm$1.5}} & 20.0{{$\pm$2.2}}& 23.5 {{$\pm$3.0}} & 8.1{{$\pm$1.2}}\\
  \bottomrule
\end{tabular}}

\end{table*}


\subsection{Dataset, Baselines \& Metrics}
\subsubsection{Datasets:}
We conduct experiments on a synthetic datasets and a benchmark face dataset CelebA.

\textbf{Synthetic}: We build two synthetic datasets which include images of causally related objects. The first one is named Pendulum. Each image contains 3 entities (\textsc{pendulum}, \textsc{light}, \textsc{shadow}), and 4 concepts ((\textsc{pendulum angle}, \textsc{light angle}) $\to$ (\textsc{shadow location}, \textsc{shadow length})). The second one is named Flow. Each image contains 4 concepts (\textsc{ball size} $\to$ \textsc{water size}, (\textsc{water size}, \textsc{hole})$\to$ \textsc{water flow}). Due to page limitation, main text only shows the  results on Pendulum, and experiments  on Flow and more details of two datasets are given in Appendix C.1. 

\textbf{Real world benchmark}: We also use a real world dataset CelebA\footnote{http://mmlab.ie.cuhk.edu.hk/projects/CelebA.html}\cite{DBLP:conf/iccv/LiuLWT15}, a widely used dataset in the computer vision community. In this dataset, there are in total 200k human face images with labels on different concepts, and we choose two subsets of causally related attributes. The first set is  CelebA(\textsc{Smile}), which  consists of \textsc{gender}, \textsc{smile},  \textsc{eyes open}, \textsc{mouth open}. The second one is CelebA(\textsc{Beard}), which consists of \textsc{age}, \textsc{gender},  \textsc{bald}, \textsc{beard}. Main text only shows  results on CelebA(\textsc{Smile}), and more experimental results on other concepts are provided in the Appendix D.

 
\textbf{Baselines:} We compare our method with some state of the arts and show the results of ablation study. Baselines are categorized into supervised and unsupervised methods. 

CausalVAE-unsup, LadderVAE \cite{laddervae} and $\beta$-VAE \cite{higgins2017beta} are unsupervised methods.   CausalVAE-unsup is a reduced version of our model whose structure is the same as CausalVAE except that the Mask Layer and the supervision conditional prior $p(\mathbf{z}|\mathbf{u})$ are removed. 

Supervised methods include disentangled representation learning method ConditionVAE \cite{conditionvae}, which does not include causal layers in the model structure and causal generative model CausalGAN \cite{causalgan}, which needs the true causal graph to be given as a prior. 

As CausalGAN  does not focus on representation learning, we only compare our CausalVAE with CausalGAN on intervention experiment (results given in Appendix D.3).  For these methods, the prior conditioning on the labels are given, and the dimensionality of the latent representation is the same as CausalVAE.

\textbf{Metrics:} We use Maximal Information Coefficient (MIC) and Total Information Coefficient (TIC) \cite{kinney2014equitability} as our evaluation metrics. Both of them indicate the degree of information relevance between the learned representation and the ground truth labels of concepts. 
\subsection{Intervention experiments}

Intervention experiments aim at testing if a certain dimension of the latent representation has interpretable semantics. The value of a latent code is manipulated  by "do-operation" as introduced in previous sections, and we observe how the generated image appears. 
\begin{figure}[h]

\begin{center}
\vskip -0.1in
\centerline{\includegraphics[width=0.5\columnwidth]{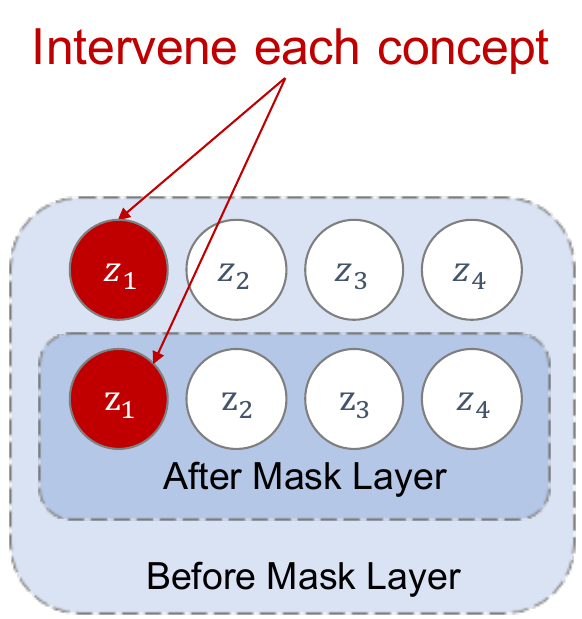}}
\caption{Intervention method}
\label{Intervention}
\vskip -0.2in
\end{center}

\end{figure}
Intervention is conducted by the following steps: 
\begin{itemize}
    \item A generative model is trained.
    \item An arbitrary image from the training set is fed to the encoder to generate a latent code $\mathbf{z}$.
    \item We manipulate the value of $z_i$ corresponding to a  concept of interest. For CausalVAE, as Fig. \ref{cvae_structure} \textcircled{4} and Fig. \ref{Intervention} show, we need to manipulate both the input and output nodes of the SCM layer. Note that the effect of manipulation to a parental node will be propagated to its children.
    \item The intervened latent code $\tilde{\mathbf{z}}$ passes through the decoder to generate a new image. In the experiments, all images in the dataset are used to train our proposed model CausalVAE and other baselines.
\end{itemize}

Hyperparameters $(\alpha,\beta,\gamma)=(1,1,1)$ for all experiments unless specified. 






We first conduct intervention experiments on the Pendulum dataset, with 4 latent concepts and results are given in Fig. \ref{pendulum_res}. We intervene a certain concept by setting the corresponding latent code value to 0. We expect that the pattern of the manipulated concept will be fixed across all images under the same intervention. For example, when we intervene the pendulum \textsc{angle} as shown in the first line of Fig. \ref{pendulum_res} (a), the \textsc{angle} of pendulum of different images are almost the same. Meanwhile, we also observe that the \textsc{Shadow location} and \textsc{Shadow length} change in a correct way that aligns with the physics law. Note that this is also related to the concept of modularity, meaning that intervening a certain part of the generative system usually does not affect the other parts of the system. Similar phenomenon is observed in other intervention experiments, demonstrating that our model correctly implements the underlying causal system. 
The results of ConditionVAE, a supervised method without considering the causal structure, are given in Fig. \ref{pendulum_res} (b). There exists a problem that manipulating the latent codes of effects sometimes has no influence to the whole image. This is probably because they do not explicitly consider causal disentanglement.

We also design another synthetic dataset Flow and do the same comparative experiments on that and the results support our claim. Because of page limitation, we show the results in Appendix D.

Fig. \ref{do_smile} demonstrates the good result of CausalVAE on real world banchmark dataset CelebA, with subfigures showing the experiments on intervening concepts \textsc{gender, smile, eyes open} and \textsc{mouth open} respectively. We observe that when we intervene the cause concept \textsc{smile}, the status of \textsc{mouth open} also changes. In contrast, intervening effect concept \textsc{mouth open} does not cause the cause concept \textsc{smile} to change. Table 1 records the mutual information (MIC/TIC) between the learned representation and the ground truth concept labels of all compared methods. Our model achieves best alignment with the concept labels, justifying the effectiveness of our proposed method. On the contrary, factors learned by those compared methods have low correlation with the ground truth labels, indicating that those factors are at least not corresponding to the causal concepts of interest. 

In addition, we show in Fig. \ref{learn_A} the learned adjacency matrix $\mathbf{A}$. To learn a precise causal graph, we design a pre-train process by optimizing augmented Lagrangian method \cite{yu2019dag} on Eq. \ref{constraint_A}, details are shown in Appendix C.3. As the training epoch increases, we see that the graph learned by our model quickly converges to the true one, which shows that our method is able to correctly learn the causal relationship among the factors.

\section{Conclusion}

In this paper, we investigate an important task of learning disentangled representations of causally related concepts in data, and propose a new framework called CausalVAE which includes a SCM layer to model the causal generation mechanism of data. We prove that the proposed model is fully identifiability given additional supervision signal. Experimental results with synthetic and real data show that CausalVAE successfully learns representations of causally related concepts and allows intervention to generate counterfactual outputs as expected according to our understanding of the causal system. To the best of our knowledge, our work is the first one that successfully implement causal disentanglement and is expected to bring new insights into the domain of disentangled representation learning.

{\small
\bibliographystyle{ieee_fullname}
\bibliography{egbib}
}

\end{document}


\title{Supplementary Material of\\ CausalVAE:
Disentangled Representation Learning\\ via Neural Structural Causal Models}
\author{Mengyue Yang$^{1,2}$, Furui Liu$^{1,}$\thanks{Corresponding author.} , Zhitang Chen$^{1}$, Xinwei Shen$^{3}$, Jianye Hao$^{1}$, Jun Wang$^{2}$\\
  $^{1}$ Noah's Ark Lab, Huawei, Shenzhen, China \\
  $^{2}$ University College London, London, United Kingdom\\
  $^{3}$ The Hong Kong University of Science and Technology, Hong Kong, China\\
  {\tt\small \{yangmengyue2,liufurui2,chenzhitang2,haojianye\}@huawei.com}\\
  {\tt\small xshenal@connect.ust.hk}\\ {\tt\small jun.wang@cs.ucl.ac.uk} 
}
\maketitle
\pagestyle{empty}  
\thispagestyle{empty} 
\appendix

\section{Proof of Proposition 1}
Write the KL term in ELBO defined in Eq. 8 in the main text as
\[
\begin{split}
&\mathcal{D}[q_{\phi}(\bm{\epsilon},\mathbf{z}|\mathbf{x},\mathbf{u})\Vert p_{\theta}(\bm{\epsilon},\mathbf{z}|\mathbf{u})] \\=&  \iint q_{\phi}(\bm{\epsilon},\mathbf{z}|\mathbf{x},\mathbf{u})\log \frac{q_{\phi}(\bm{\epsilon},\mathbf{z}|\mathbf{x},\mathbf{u})}{p_{\bm{\epsilon}}(\bm{\epsilon})p_{\theta}(\mathbf{z}|\mathbf{u})}d\bm{\epsilon}d\mathbf{z}\\
=&\iint q_{\phi}(\bm{\epsilon},\mathbf{z}|\mathbf{x},\mathbf{u})\log \frac{q_{\phi}(\bm{\epsilon},\mathbf{z}|\mathbf{x},\mathbf{u})}{p_{\bm{\epsilon}}(\bm{\epsilon})}d\bm{\epsilon}d\mathbf{z}\nonumber\\ 
&+ \iint q_{\phi}(\bm{\epsilon},\mathbf{z}|\mathbf{x},\mathbf{u})\log \frac{q_{\phi}(\bm{\epsilon},\mathbf{z}|\mathbf{x},\mathbf{u})}{p_{\theta}(\mathbf{z}|\mathbf{u})}d\bm{\epsilon}d\mathbf{z}\\
&- \iint q_{\phi}(\bm{\epsilon},\mathbf{z}|\mathbf{x},\mathbf{u})\log q_{\phi}(\bm{\epsilon},\mathbf{z}|\mathbf{x},\mathbf{u})d\bm{\epsilon}d\mathbf{z},
\end{split}
\]
The third term in above equation could be rewritten as a constant. Details are shown as below.
\begin{align}
&- \iint q_{\phi}(\bm{\epsilon},\mathbf{z}|\mathbf{x},\mathbf{u})\log q_{\phi}(\bm{\epsilon},\mathbf{z}|\mathbf{x},\mathbf{u})d\bm{\epsilon}d\mathbf{z}\nonumber\\
=&-\iint q(\epsilon|x,u)\delta(z=C\epsilon)\log q(\epsilon|x,u)d\epsilon dz \nonumber\\&- \iint q(\epsilon|x,u)\delta(z=C\epsilon)\log\delta(z=C\epsilon) d\epsilon dz\nonumber\\
=&H(q_\phi(\epsilon|x,u))-0=H(\mathcal{N}(\mu_\phi(x,u),s\mathrm{I})))\nonumber\\
=& const,
\end{align}
In our method, we ignore this term in ELBO expression.
Then, based on Eq. 9 in the main text, we have
\[
\begin{split}
&\iint q_{\phi}(\bm{\epsilon},\mathbf{z}|\mathbf{x},\mathbf{u})\log \frac{q_{\phi}(\bm{\epsilon},\mathbf{z}|\mathbf{x},\mathbf{u})}{p_{\bm{\epsilon}}(\bm{\epsilon})}d\bm{\epsilon}d\mathbf{z}\\
=&\int q_{\phi}(\bm{\epsilon}|\mathbf{x},\mathbf{u})\log \frac{q_{\phi}(\bm{\epsilon}|\mathbf{x},\mathbf{u})}{p_{\bm{\epsilon}}(\bm{\epsilon})}\int\delta(\mathbf{z}=\mathbf{C}\bm{\epsilon})d\mathbf{z}d\bm{\epsilon}\\
&+ \int q_{\phi}(\bm{\epsilon}|\mathbf{x},\mathbf{u})\int \delta(\mathbf{z}=\mathbf{C}\bm{\epsilon})\log \delta(\mathbf{z}=\mathbf{C}\bm{\epsilon}))d\mathbf{z} d\bm{\epsilon}\\
=&\mathcal{D}[q_{\phi}(\bm{\epsilon}|\mathbf{x},\mathbf{u})\Vert p_{\bm{\epsilon}}(\bm{\epsilon})]+0\\
=&\mathcal{D}[q_{\phi}(\bm{\epsilon}|\mathbf{x},\mathbf{u})\Vert p_{\bm{\epsilon}}(\bm{\epsilon})],
\end{split}
\]
and
\[
\begin{split}
&\iint q_{\phi}(\bm{\epsilon},\mathbf{z}|\mathbf{x},\mathbf{u})\log \frac{q_{\phi}(\bm{\epsilon},\mathbf{z}|\mathbf{x},\mathbf{u})}{p_{\theta}(\mathbf{z}|\mathbf{u})}d\bm{\epsilon}d\mathbf{z}\\
=&\int q_{\phi}(\bm{z}|\mathbf{x},\mathbf{u})\log \frac{q_{\phi}(\bm{z}|\mathbf{x},\mathbf{u})}{p_{\theta}(\mathbf{z}|\mathbf{u})}\int\delta(\bm{\epsilon}=\mathbf{C}^{-1}\bm{z})d\bm{\epsilon}d\mathbf{z}\\
&+ \int q_{\phi}(\bm{z}|\mathbf{x},\mathbf{u})\int \delta(\bm{\epsilon}=\mathbf{C}\bm{z})\log \delta(\bm{\epsilon}=\mathbf{C}^{-1}\bm{z}) d\bm{\epsilon} d\mathbf{z}\\
=&\mathcal{D}[q_{\phi}(\bm{z}|\mathbf{x},\mathbf{u})\Vert p_{\theta}(\mathbf{z}|\mathbf{u})]+0\\
=&\mathcal{D}[q_{\phi}(\bm{z}|\mathbf{x},\mathbf{u})\Vert p_{\theta}(\mathbf{z}|\mathbf{u})].
\end{split}
\]
Adding up the above two terms leads to the desired form of Proposition 1.




\section{Identifiability}
\subsection{Proof of Theorem 1}\label{identifiable1}
The general logic of the proofing follows \cite{nonlinearica}, but we focus on both encoder and decoder. In our setting, we has joint latent variables $\bm{\epsilon}, \mathbf{z}$, and we prove identidfiabilty of both of them. 

Another different setting from iVAE is that we consider a slighter transformation matrix, since our additional observations $\mathbf{u}$ of each concepts align to each causal representations $\mathbf{z}$.

\textbf{Sketch of proof:}  

We analyze the identifiability of $\bm{\epsilon}$ starting with $p_{\bm{{\bm{{\bm{\theta}}}}}}\mathbf{(\mathbf{x}|u)}=p_{\widetilde{{\bm{\theta}}}}(\mathbf{x}|\mathbf{u})$. Then we define a new invertible matrix $\mathbf{L}$ which contains additional observation $u_i$ in causal system, and use it to prove that the learned $\mathbf{\tilde{T}}$ is the transformation of $\mathbf{T}$.  
{Step 2:}
We take the inference model into consideration and analyze the identifiablity of the inference model by relating the inference model to the generative model.

\textbf{Details:}  

At the begining of proof, we consider a simple condition that the dimension of observation data $d$ equals to the dimension of latent variables $n$.

The distribution has two sufficient statistics, the mean and variance of $\mathbf{z}$, which are denoted by sufficient statistics $\mathbf{T}(\mathbf{z}) = (\bm{\mu}(\mathbf{z}), \bm{\sigma}(\mathbf{z}))=(T_{1,1}(z_1),\dots,T_{n,2}(z_n))$. We use these notations for model identifiability analysis in Section 5. To simplify proof process, we absorb the injective functions $\mathbf{g}(\cdot)$ into generate model $\mathbf{f}(\cdot)$ since mask layer will not influence the quality of disentangled representation $\mathbf{z}$.

\begin{align}
    &p_{\bm{\theta}}\mathbf{(\mathbf{x}|u)}=p_{\widetilde{{\bm{\theta}}}}(\mathbf{x}|\mathbf{u}), \nonumber \\
    \Rightarrow	&\iint_{\mathbf{z},\bm{\epsilon}}  p_{\bm{\theta}}(\mathbf{x}|\mathbf{z},\bm{\epsilon})p_{{\bm{\theta}}}(\mathbf{z},\bm{\epsilon}|\mathbf{u})d\mathbf{z}d\bm{\epsilon} \nonumber\\ 
    &= \iint_{\mathbf{z},\bm{\epsilon}}  p_{\widetilde{{\bm{\theta}}}}(\mathbf{x}|\mathbf{z},\bm{\epsilon})p_{\widetilde{{\bm{\theta}}}}(\mathbf{z},\bm{\epsilon}|\mathbf{u})d\mathbf{z}d\bm{\epsilon},\nonumber \\
    \Rightarrow &\int_{\mathbf{z}}  p_{\bm{\theta}}(\mathbf{x}|\mathbf{z})p_{{\bm{\theta}}}(\mathbf{z}|\mathbf{u})d\mathbf{z} = \iint_{\mathbf{z}}  p_{\widetilde{{\bm{\theta}}}}(\mathbf{x}|\mathbf{z})p_{\widetilde{{\bm{\theta}}}}(\mathbf{z}|\mathbf{u})d\mathbf{z},\nonumber \\
    \Rightarrow	&\int_{\mathbf{x}'} p_{{\bm{\theta}}}(\mathbf{x}|\mathbf{f^{-1}}(\mathbf{x}'))p_{{\bm{\theta}}}(\mathbf{f^{-1}}(\mathbf{x}')|\mathbf{u})|\det(J_{\mathbf{f^{-1}}}(\mathbf{x}'))|d\mathbf{x}'\nonumber\\
    &= \int_{\mathbf{x}'} p_{{\bm{\theta}}}(\mathbf{x}|\mathbf{\tilde{f}^{-1}}(\mathbf{x}'))p_{\widetilde{{\bm{\theta}}}}(\mathbf{\tilde{f}^{-1}}(\mathbf{x}')|\mathbf{u})|\det(J_{\mathbf{\tilde{f}^{-1}}}(\mathbf{x}'))|d\mathbf{x}'. \label{eq1}
\end{align}

In determining function $\mathbf{f}$, there exist a Gaussian distribution $p_{\bm{\xi}}(\bm{\xi})$ which has infinitesimal variance. 
Then, the $p_{{\bm{\theta}}}(\mathbf{x}|\mathbf{f^{-1}}(\mathbf{x}'))$ can be written as $p_{\bm{\xi}}(\mathbf{x}-\mathbf{x}')$. As the assumption (1) holds, this term is vanished. Then in our method, there exists the following equation:
\begin{align}
     p_{{\bm{\theta}}}(\mathbf{f^{-1}}(\mathbf{x}')|\mathbf{u})|\det(J_{\mathbf{f^{-1}}}(\mathbf{x}'))|&= p_{\widetilde{{\bm{\theta}}}}(\mathbf{\tilde{f}^{-1}}(\mathbf{x}')|\mathbf{u})|\det(J_{\widetilde{\mathbf{f}}^{-1}}(\mathbf{x}'))|,\nonumber\\
    \Rightarrow\widetilde{p}_{\bm{\theta}}(\mathbf{x}) &= \widetilde{p}_{\widetilde{{\bm{\theta}}}} (\mathbf{x}).
\end{align}

Adopting the definition of multivariate Gaussian distribution,  we define 
\begin{equation}
    \bm{\lambda}_s(\mathbf{u})= \left[
        \begin{array}{ccc}
          \lambda_1^s(u_1) &  &  \\
            & \ddots &  \\
            &  & \lambda_n^s(u_n) \\
         \end{array}\right].
\end{equation}

There exists the following equations:
\begin{align}
    &\log|\det(J_{\mathbf{f}^{-1}}(\mathbf{x}))| - \log \mathbf{Q}(\mathbf{f}^{-1}(\mathbf{x}))+\log\mathbf{Z(u)}\\&+\sum_{s=1}^2 \mathbf{T}_s (\mathbf{f}^{-1}(\mathbf{x}))\bm{\lambda}_s(\mathbf{u}),\nonumber\\
    = &\log|\det(J_{\widetilde{\mathbf{h}}}(\mathbf{x}))| -\log \widetilde{\mathbf{Q}}(\tilde{\mathbf{f}}^{-1}(\mathbf{x}))+\log\mathbf{\tilde{Z}(u)}\nonumber\\&+\sum_{s=1}^2\widetilde{\mathbf{T}}_{s}(\tilde{\mathbf{f}}^{-1}(\mathbf{x}))\widetilde{\bm{\lambda}}_s(\mathbf{u}),\label{original}
\end{align}
where $\mathbf{Q}$ denotes the base measure. In Gaussian distribution, it is $\bm{\sigma}(\mathbf{z})$.

In learning process, $\mathbf{\widetilde{A}}$ is restricted as DAG. Thus, the $\mathbf{\widetilde{C}}$ exists which is full rank matrix. The item which is not related to $u$ in Eq. \ref{original} are cancelled out \cite{sorrenson2020disentanglement}. 
\begin{align}
    &\sum_{s=1}^2{\mathbf{T}}_{s}(\mathbf{f}^{-1}(\mathbf{x}))\bm{\lambda}_s(\mathbf{u})=
    \sum_{s=1}^2\widetilde{\mathbf{T}}_{s}(\tilde{\mathbf{f}}^{-1}(\mathbf{x}))\widetilde{\bm{\lambda}}_s(\mathbf{u}) + \mathbf{b}, \label{eqsimple2}
\end{align}
where $\mathbf{b}$ is a vector related to $\mathbf{u}$.

In our model, there exist a deterministic relationship $\mathbf{C}$ between $\bm{\epsilon}$ and $\mathbf{z}$ where $\mathbf{C} = (\mathbf{I}-\mathbf{A}^T)^{-1}$. Thus we could get equivalent of Eq. \ref{eqsimple2} as follows,
\begin{align}
    &\sum_{s=1}^2{\mathbf{T}}_{s}(\mathbf{C}\mathbf{h}(\mathbf{x}))\bm{\lambda}_s(\mathbf{u})=
    \sum_{s=1}^2\widetilde{\mathbf{T}}_{s}(\widetilde{\mathbf{C}}\widetilde{\mathbf{h}}(\mathbf{x}))\widetilde{\bm{\lambda}}_s(\mathbf{u}) + \mathbf{b}‘, \label{eqsimple3}
\end{align}
where $s$ denote the index of sufficient statistics of  Gaussian distributions, indexing the mean (1) and the variance (2).

By assuming that the additional observation $u_i$ is  different, it is guaranteed that coefficients of the observations for different concepts are distinct. Thus, there exists an invertible matrix corresponding to additional information $\mathbf{u}$:
\begin{align}
    \mathbf{L}= \left[
    \begin{array}{ccc}
      \bm{\lambda}_1(\mathbf{u}) &  \\
        & \bm{\lambda}_2(\mathbf{u})
     \end{array}\right].\label{matl}
\end{align}
Since the assumption that $u_i\not=0$ holds, $\mathbf{L}$ is $2n \times 2n$ invertible and full rank diagonal matrix. Then, function of $\bm{\lambda}$ in Eq. \ref{eqsimple2} and Eq. \ref{eqsimple3} are replcaed by Eq. \ref{matl}, we could get:
\begin{align}
    &\mathbf{LT}(\mathbf{f}^{-1}(\mathbf{x})) = \widetilde{\mathbf{L}}\widetilde{\mathbf{T}}(\widetilde{\mathbf{f}}^{-1}(\mathbf{x}))+ \mathbf{b},\\
    &\mathbf{T}(\mathbf{f}^{-1}(\mathbf{x})) = \mathbf{B_2}\widetilde{\mathbf{T}} (\widetilde{\mathbf{f}}^{-1}(\mathbf{x}))  + \mathbf{b_2},\label{iz}
\end{align}
where 
\begin{equation}
    \mathbf{B_2}= \left[
        \begin{array}{ccc}
          {\lambda}_{1,1}(u_1)^{-1}\widetilde{\lambda}_{1,1}(u_1) &  &  \\
            & \ddots &  \\
            &  & {\lambda}_{n,2}(u_n)\widetilde{\lambda}_{n,2}(u_n) \\
         \end{array}\right].
\end{equation}
We replace $\mathbf{f}^{-1}$ with $\mathbf{Ch}$ and we could get the equations as below:
\begin{align}
    \mathbf{L}\mathbf{T}(\mathbf{Ch}(\mathbf{x})) = \widetilde{\mathbf{L}}{\widetilde{\mathbf{T}}}(\tilde{\mathbf{C}}\tilde{\mathbf{h}}(\mathbf{x}))
    \Rightarrow \mathbf{T}(\mathbf{h}(\mathbf{x})) = \mathbf{B_1}{\widetilde{\mathbf{T}}} (\widetilde{\mathbf{h}}(\mathbf{x})) + \mathbf{b_1},\label{ie}
\end{align}
where $\mathbf{B_3}=\mathbf{C\tilde{C}^{-1}}$ is invertible matrix which corresponds to $\mathbf{C}$ and $\mathbf{B_1=\mathbf{L^{-1}B_3^{-1}\widetilde{L}}}$. The definition of $\widetilde{\mathbf{L}}$ on learning model migrates the definition of $\mathbf{L}$ on ground truth.

Then we adopt the definitions following \cite{nonlinearica}. According to the Lemma 3 in \cite{nonlinearica}, we are able to pick out a pair $(\epsilon_i,\epsilon_i^2)$ such that, $(\textbf{T}'_i(z_i),\textbf{T}'_i(z_i^2))$ are linearly independent. Then concat the two points into a vector, and denote the Jacobian matrix $\mathbf{Q} = [J_\mathbf{T}(\bm{\epsilon}),J_\mathbf{T}(\bm{\epsilon}^2)]$, and   define $\mathbf{\tilde{Q}}$ on $\mathbf{\tilde{T}}(\mathbf{\tilde{h}} \circ \mathbf{Cf}(\bm{\epsilon}))$ in the same manner. By  differentiating Eq. \ref{ie}, we get
\begin{align}
    \mathbf{Q} = \mathbf{B_1} \mathbf{\tilde{Q}}.\label{invertibleb}
\end{align}
Since the assumptiom (2) that Jacobian of $\mathbf{f}^{-1}$ is full rank holds, it can prove that both $\mathbf{Q}$ and $\mathbf{\tilde{Q}}$ are invertible matrix. Thus from Eq. \ref{invertibleb}, $\mathbf{B_1}$ is invertible matrix. Using the same way as shown in Eq. \ref{invertibleb}, it can prove that $\mathbf{B_2}$ is invertible matrix.

Eq. \ref{iz} and Eq. \ref{ie} both hold. Combining the two results supports the identifiability result in CausalVAE.


\subsection{Extension of Definition 1}
In most of scenarios, latent variable is a low dimensional representation of the observation, since we are not interested in all the information in observations. 

Therefore, we usually have $d>n$. We called it the reduction of dimension. We  add auxiliary term as $\mathbf{\lambda(x) = \{\lambda(u),\lambda'\}}$ In our model, Only $n$ components of the latent variable are
modulated, and its density has the form:
\begin{align}
    p_{\bm{\theta}}(\mathbf{z}|\mathbf{u}) = \frac{\mathbf{Q}(\mathbf{z})}{\mathbf{Z(u)}}\exp{{\sum_i^n\mathbf{T_i}(z_i)\lambda_i(u_i)}}\label{lowdimensional}
\end{align}
and the term $e^{\sum_{n+1}^d\mathbf{T}(z_i)\lambda_i}$ is simply absorbed into $\mathbf{Q(z)}$. When we evaluate Eq. \ref{original} by new definition (Eq. \ref{lowdimensional}), the dimension of $p(\mathbf{z|u})$ is $n$, because the remaining part is cancelled out.

Assume that $p_{\bm{\theta}}\mathbf{(\mathbf{x}|u)}$  equal to $p_{\widetilde{{\bm{\theta}}}}(\mathbf{x}|\mathbf{u})$. For all the observational pairs $(\mathbf{x},\mathbf{u})$, let $J_h$ denote the Jacobian matrix of the encoder function. 
Following the definition in Theorem 2 in i VAE \cite{nonlinearica}, $\mathbf{B}$ will be indexed by 4 indicates $(i,l, a,b)$, where $1<i<d$ and $1<l<s$ refer to the rows and $1<a<d$ and $1<b<s$ refer to the columns. We define a following equation:
\begin{align}
    \mathbf{v} = \mathbf{\tilde{C}\circ \tilde{h}\circ f(z)}.
\end{align}
The goal is to show  that $v_i(\mathbf{z})$ is a function of only one $z_j$. We denote by $v_i^r := \frac{\partial v_i}{\partial z_r}$ and $v_i^{rt} := \frac{\partial^2 v_i}{\partial z_r \partial z_t}$. By  differentiating Eq. \ref{iz} with respect to $z_s$, we could get:
\begin{align}
    T'_{i,l}(z_i) = \sum_{a =1}^d\sum_{b = 1}^s B_{2,(i,l,a,b)} \tilde{T}'_{a,b}(v_a(\mathbf{z}))v_a^r(\mathbf{z}).
\end{align}

\begin{lemma}
(from Lemma 9 in Khemakhem \etal~ \cite{khemakhem2020ice}):  Consider a distribution that follows a strongly exponential family. Its sufficient statistic $\mathbf{\tilde{T}}$ is differentiable almost surely. Then $\tilde{T}'_i \not= 0$ almost everywhere on $\mathbb{R}$ for all $1 \le i \le s$.
\end{lemma} 

For $r>n$, $T'_{i,l}(z_i) = 0$, according to Lemma 1, $\tilde{T}'_{a,b}(v_a(\mathbf{z})) \not= 0$, since  $\mathbf{B}_2$ is an invertible matrix, we can conclude that  $v_a^r(\mathbf{z})=0$ for all $a<n$ and $r>n$. Therefore, we can conclude that each of the first $n$ components of $\mathbf{v}$ is only a function of one different $z_j$. Thus, when $d>n$, we could get the same conclusion as Theorem 1.

\subsection{Identifiability of Causal Graph}
Consider the  identifiability analysis in Appendix \ref{identifiable1}. For the framework of CausalVAE, in Causal Layer, the latent code $\mathbf{z}$ is identified since $\mathbf{B_2}$ is a diagonal matrix which corresponds to learnt $\tilde{\mathbf{z}}$ and $\mathbf{z}$. Since the true $\bm{\epsilon}$ and learnt $\tilde{\bm{\epsilon}}$ are linearly related, $\mathbf{B_1}$, $\mathbf{C}$ and $\tilde{\mathbf{C}}$ are in a linear equivalent class. In other words,  $\mathbf{C}$ or $\mathbf{A}$ is identifiable in Causal Layer up to a linear equivalent class.

In our work, strict identifiability is guaranteed by the non-linear mask layer. Details of the Mask Layer are shown in Section 3.2 in main text. The Mask Layer uses non-linear functions and additional supervision signal $\mathbf{u}$ (non-Gaussian) to help the model to identify the true causal graph in a linear equivalent class.

\section{Implementation Details}
We use one NVIDIA Tesla P40 GPU as our training and inference device.

For the implementation of CausalVAE and other baselines, we extend $\mathbf{z}$ to matrix $\mathbf{z}\in\mathbb{R}^{n\times k}$ where $n$ is the number of concepts and $k$ is the latent dimension of each $\mathbf{z}_i$. The corresponding prior or conditional prior distributions of CausalVAE and other baselines are also adjusted (this means that we extend the multivariate Gaussian to the matrix Gaussian). 

The subdimensions $k $ for each synthetic (pendulum, water)   experiments are set to be 4, and 32 for CelebA experiments. The implementation of continuous DAG constraint $H(\mathbf{A})$ follows the code of \cite{yu2019dag} \footnote{https://github.com/fishmoon1234/DAG-GNN}.

\subsection{Data Preprocessing}
\subsubsection{Sythetic Simulator}
\begin{figure}[ht]
\begin{center}
\centerline{\includegraphics[width=0.45\columnwidth]{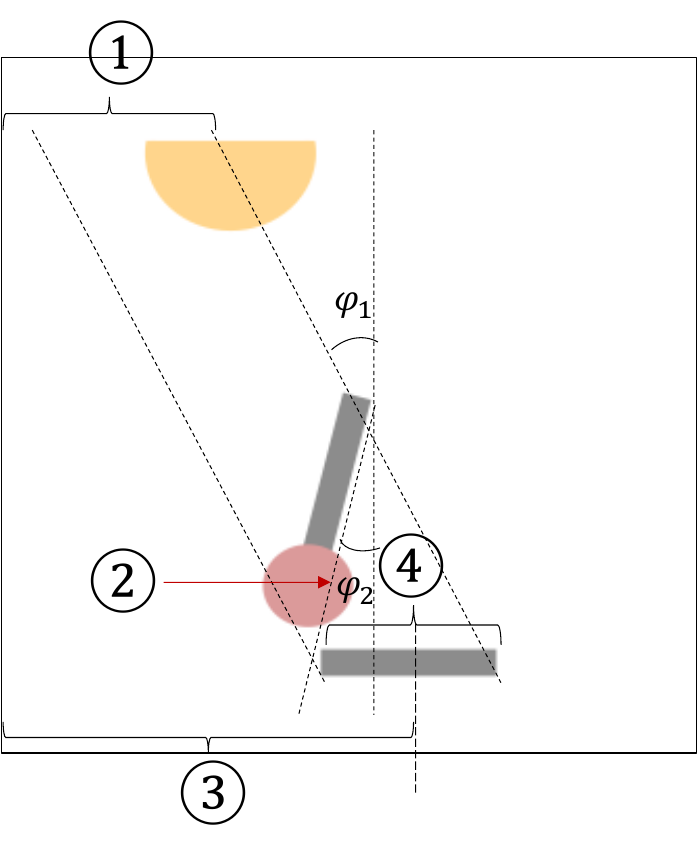}}
\caption{Generate Policy of Pendulum Simulator}
\label{pendulum_simulator}
\end{center}
\end{figure}

\begin{figure}[ht]
\begin{center}
\centerline{\includegraphics[width=0.5\columnwidth]{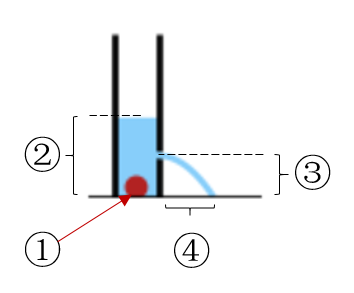}}
\caption{Generate Policy of Flow Simulator}
\label{flow_simulator}
\end{center}
\end{figure}
Fig. \ref{pendulum_simulator} shows our policy of generating synthetic Pendulum data. The picture includes a pendulum. The angles of pendulum and the light are changing overtime, and  projection laws are used to generate the shadows. Given the light \textsc{position} and pendulum \textsc{angle}, we  get the angles $\varphi_1$ and $\varphi_2$. Then the system can calculate the shadow \textsc{position} and \textsc{length} using triangular functions. The causal graph of concepts is shown in Fig. \ref{causal_example} (a). In Pendulum generator, the image size is set to be $96 \times 96$ with 4 channels. We generate about 7k images (6k for training and 1k for inference), $\varphi_1$ and $\varphi_2$ are ranged in around $[-\frac{\pi}{4},\frac{\pi}{4}]$, and they are generated independently. For each image, we provide 4 labels, which include light position, pendulum angle, shadow position and shadow length. For light position, we use the value of center of semicircle (Fig.\ref{pendulum_simulator} \textcircled{1}) as supervision signal. For the pendulum angle, we use the value of $\phi_2$ as supervision signal (Fig. \ref{pendulum_simulator} \textcircled{2}). For shadow position and shadow length, we use the length of Fig.\ref{pendulum_simulator} \textcircled{3} and Fig.\ref{pendulum_simulator} \textcircled{4} as supervision signal respectively.

Fig. \ref{flow_simulator} presents our policy of generating synthetic Flow data. Each image is of the $96 \times 96\times4$ resolution, and consists of a cup of water and a ball. The original water level, the ball size (Fig.\ref{flow_simulator}\textcircled{1}) and the location of hole (Fig.\ref{flow_simulator}\textcircled{3}) vary over time. 
Given the ball size Fig. \ref{flow_simulator} and the original water level, we determine the \textsc{water height} (Fig.\ref{flow_simulator}\textcircled{2}). Then we generate \textsc{water flow} according to the Parabola law, where we additionally introduce a noise from $\mathcal{N}(0, 0.01)$ to the gravitational acceleration. The causal graph of concepts is given in Fig. \ref{causal_example} (b).
We consider four semantically meaningful concepts, \textsc{balls size}, \textsc{water height}, \textsc{hole position} and \textsc{water flow}, whose supervised signals are given by the ball's diameter (Fig.\ref{pendulum_simulator} \textcircled{1}), the length of Fig. \ref{pendulum_simulator} \textcircled{2}, the length of Fig.\ref{pendulum_simulator} \textcircled{3} and Fig.\ref{pendulum_simulator} \textcircled{4} respectively. The sample size is 8k with 6k for training and 2k for testing.

\subsubsection{Data Preprocess of CelebA}
CelebA dataset contains 20K human face images. We preprocess the original dataset by following two steps:

(1) We divided the whole dataset into training dataset $85\%$ and test dataset $15\%$.

(2) We  only focus on  facial features and resize the picture to be squared ($128\times128$ with 3 channels).

\subsection{Intervention Experiments}

\subsubsection{Synthetic}
In synthetic experiments, we train the model on synthetic data for 80 epochs, and use this model to generate latent code of representations. The hyperparameters of baselines are defined as default.

For CausalVAE, we set the $\alpha = 0.3$ and $(\beta, \gamma) = (1,1)$. We use $\mathcal{N}(\mathbf{u,|u|})$ as the condition prior $p_{\bm{\theta}}(\mathbf{z|u})$. In the implementation of CausalVAE, $|\mathbf{z}_{\text{mean}}|$ is used as the variance of condition prior.




The details of the neural networks are shown in Table \ref{do_table1}. We all follows the neural network design strategy of Khemakhem \etal~ \cite{khemakhem2020ice} to satisfy Theorem 1 assumption (ii).

\subsubsection{CelebA}

We also present the DO-experiments of CausalVAE and CausalGAN. In the training of the models, we use face labels (\textsc{age}, \textsc{gender} and \textsc{beard}). 

For CausalVAE, we set the $\alpha = 0.3$ and $(\beta, \gamma) = (1,1)$. We use $\mathcal{N}(\mathbf{u,I})$ as the condition prior $p_{\bm{\theta}}(\mathbf{z|u})$. For all the baseline, default hyperparameters and one common encoder and decoder structure are employed. For CausalGAN, we use the publicly available code\footnote{https://github.com/mkocaoglu/CausalGAN}. 

For all the VAE-based methods, mean and variance of the distribution of the latent variable are learned during training, and the latent code $z$ are sampled from Conditional Gaussian Distribution $p_{\bm{\theta}}(\mathbf{z|u})$. In all experiments, we rescale the variance of learned representation $\mathbf{z}$ by multiplying a factor 0.1 to the original one.

Training epoches for the model  is set to be 80, and our proposed CausalVAE has a pretrain step to learn causal graph $\mathbf{A}$, which takes 10 epochs.

The details of the neural networks are shown in Table \ref{do_table2}.

\begin{figure*}[h]

\begin{center}
\centerline{\includegraphics[width=2\columnwidth]{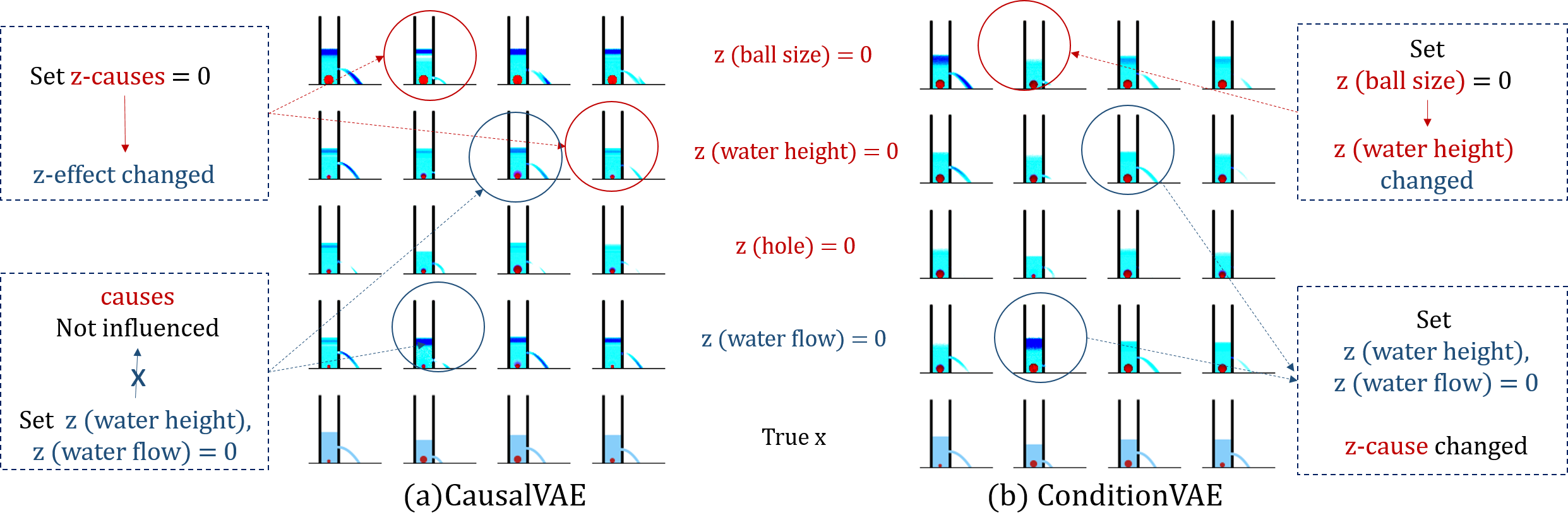}}

\caption{The results of Intervention experiments on the Flow dataset. Each row shows the result of controlling the \textsc{ball size}, \textsc{water height}, \textsc{hole}, and  \textsc{water flow} respectively. The bottom row is the original input image.}
\label{flow_res}
\end{center}

\end{figure*}

\subsection{The Pretrain Step for  Causal Graph Learning}
In our model, we need to learn the latent representation $\mathbf{z}$ and causal graph $\mathbf{A}$ simultaneously, whose optimal solution is not easy to find. Thus we adopt a pretrain stage to learn the causal graph $\mathbf{A}$ in the Mask Layer. We adopt the augmented Lagrangian to learn $\mathbf{A}$ in CausalVAE from the labels $\mathbf{u}$ in Mask Layer first. During the pretrain process, we truncate the gradient of other part of model and solve the optimization problem in Eq. \ref{pretrain_eq} to learn $\mathbf{A}$. 

The augmentation approach is widely used in causal discovery method, like NOTEARS \cite{shimizu2006linear}, DAG-GNN \cite{yu2019dag}. The pretrain is a stage that learns the graph by optimizing the following objective functions:
\begin{align}
    \text{minimize} & ~~~ l_u=\mathbb{E}_{q_D}\Vert \mathbf{u} -\mathbf{A}^T \mathbf{u}\Vert_2^2 \nonumber \\
    \text{subject to} & ~~~ H(\mathbf{A})=0
\end{align}
Then, we define an augmented Lagrangian:
\begin{align}\label{pretrain_eq}
    l_{pre} = l_u + \lambda H(\mathbf{A}) + \frac{c}{2} H^2(\mathbf{A})
\end{align}
where $\lambda$ is the Lagrangian multiplier and $c$ is the penalty. 

The following policy is used to update the $\lambda$ and $c$:
\begin{align}
    \lambda_{s+1} = \lambda_{s} +c_{s} H(\mathbf{A}_s)
\end{align}
$$ c_{s+1}=\left\{
\begin{aligned}
c_{s} & =  \eta c_{s}, &  ~~~ if |H(\mathbf{A}_s)| > \gamma|H(\mathbf{A}_{s-1})| \\
c_{s} & =  c_{s}, & ~~~ otherwise
\end{aligned}
\right.
$$
where $s$ is the iteration. In our experiments, we set $\eta = 10$ and $\gamma = \frac{1}{4}$.

\begin{table*}
\centering

  \label{tab:freq1}
  \begin{tabular}{cc}
    \toprule
    encoder & decoder \\
    \midrule
    4*96*96$\times$900 fc. 1ELU &   concepts$\times$( 4$\times$ 300 fc. 1ELU )\\
    900$\times$300 fc.  1ELU &  concepts$\times$ (300$\times$300 fc.   1ELU)\\
    300$\times$2*concepts*k fc.   &  concepts$\times$(300$\times$ 1024 fc. 1ELU)\\ 
    - &concepts$\times$(1024$\times$ 4*96*96 fc.) \\
    \bottomrule
\end{tabular}

\caption{Network design of models trained on synthetic data.}
\label{do_table1}
\end{table*}

\begin{table*}
\centering

  \label{tab:freq1}
  \begin{tabular}{cc}
    \toprule
    encoder & decoder\\
    \midrule
    - &  (1$\times$1 conv. 128 1LReLU(0.2), stride 1)\\
     4$\times$4 conv. 32 1LReLU (0.2), stride 2 & (4$\times$4 convtranspose. 64 1LReLU (0.2), stride 1)\\
    4$\times$4 conv. 64 1LReLU (0.2), stride 2 &  (4$\times$4 convtranspose. 64 1LReLU (0.2), stride 2)\\ 
    4$\times$4 conv. 64 1LReLU(0.2), stride 2  & (4$\times$4 convtranspose. 32 1LReLU (0.2), stride 2)\\
    4$\times$4 conv. 64 1LReLU (0.2), stride 2  & (4$\times$4 convtranspose. 32 1LReLU (0.2), stride 2)\\
    4$\times$4 conv. 256 1LReLU (0.2), stride 2 & (4$\times$4 convtranspose. 32 1LReLU (0.2), stride 2)\\
    1$\times$1 conv. 3, stride 1 & (4$\times$4 convtranspose. 3 , stride 2) \\
    \bottomrule
\end{tabular}

\caption{Network design of models trained on CelebA.}
\label{do_table2}
\end{table*}

\begin{figure*}[h]
\begin{center}
\centerline{\includegraphics[width=2\columnwidth]{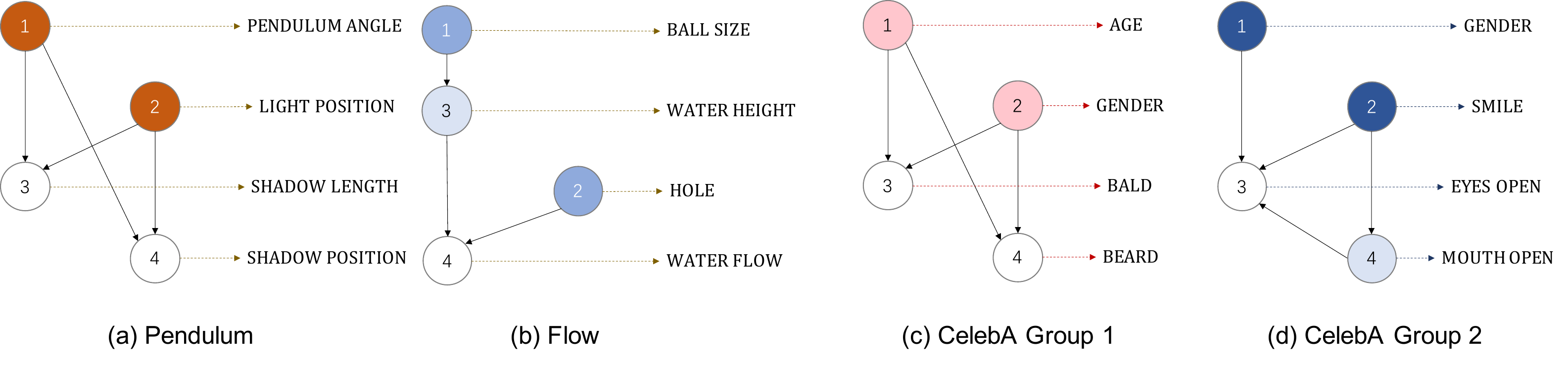}}
\caption{Causal graphs of three dataset. (a) shows the causal graph in pendulum dataset. The concepts are \textsc{pendulum} \textsc{angle}, light \textsc{position}, \textsc{shadow position} and \textsc{shadow length}. (b) shows the causal graph in CelebA, on concepts \textsc{age}, \textsc{gender} and \textsc{beard} and \textsc{bald}. (c) shows the causal graph in CelebA, on concepts \textsc{gender}, \textsc{smile}, \textsc{eyes open} and \textsc{mouth open}. }
\label{causal_example}
\end{center}
\end{figure*}

\section{Additional Experimental Results}
In this section, we show more experimental results.  Fig. \ref{causal_example} shows the causal graph among concepts in different dataset respectively.  We here show results including experiments analyzing the properties of learned representation, intervening results and the learning process of the  causal graph.

\subsection{The Property of Learned Representation}
We test our method and baselines on both synthetic data and benchmark human face data. In the previous section, we already show the relationships between the learned representation $\tilde{\mathbf{z}}$ and the target representation $\mathbf{z}$  (related by a linear transformation formed as a diagonal matrix). In this section, we visualize it by scatter plot.  

One of the important aspect of the generative model is that whether the learned representation aligns to the conditional prior we set. Our conditional prior is generated by the true label of each concept. The results show that the learned representations align to the expected representations. In figures, points are sampled from the joint  distribution, and each color corresponds to one dimension. 


The additional observations (labels) of Pendulum dataset and those of  CelebA dataset are different. In Pendulum, the labels are  values within a fixed range The labels in CelebA dataset are discrete (in $\{-1,1\}$). Thus the scatter plots are different. 

The results show that the performance of our proposed method is better than all the baselines, including the supervised method and unsupervised method. 
\begin{figure*}[h]

\begin{center}
\centerline{\includegraphics[width=2\columnwidth]{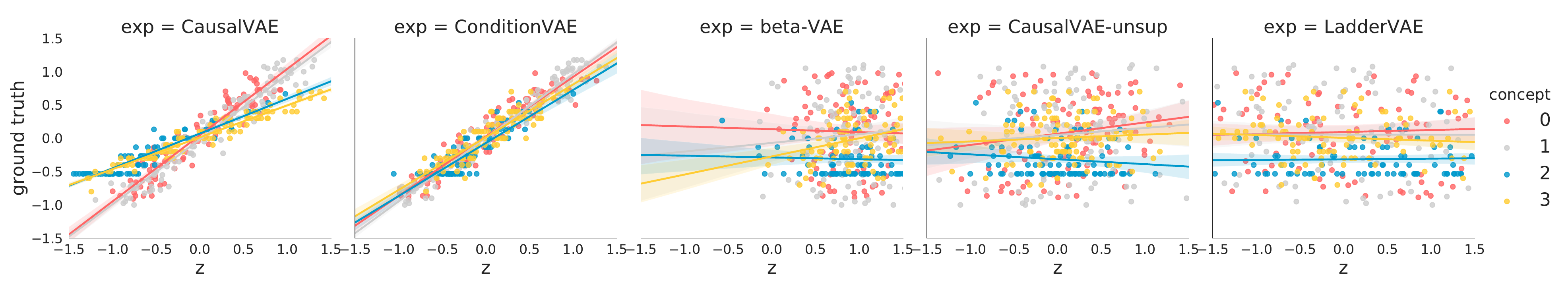}}
\caption{The figure shows the alignment of ground truth $p(\mathbf{z}|\mathbf{u})$ and the learned latent factors $q(\mathbf{z}|\mathbf{x,u})$  on pendulum experiments. Although ConditionVAE is also the supervised method, our proposed CausalVAE shows a better performance.  }
\label{align_pendulum}
\end{center}

\end{figure*}

\begin{figure*}[h]

\begin{center}
\centerline{\includegraphics[width=2\columnwidth]{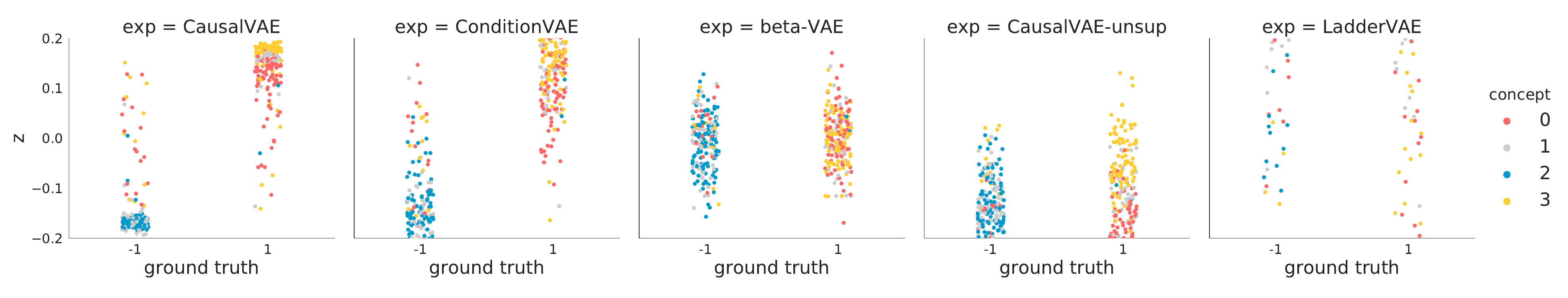}}
\caption{The figure shows the alignment of ground truth $p(\mathbf{z}|\mathbf{u})$ and the learned latent factors $q(\mathbf{z}|\mathbf{x,u})$ on CelebA for the concepts (\textsc{Beard}). The ground truth is a discrete distribution over $\{-1,1\}$, and the color of the points indicates different dimensions. The factors learned by CausalVAE show the best alignment among all.}
\label{align_beard}
\end{center}

\end{figure*}

\begin{figure*}[h]
\begin{center}
\centerline{\includegraphics[width=2\columnwidth]{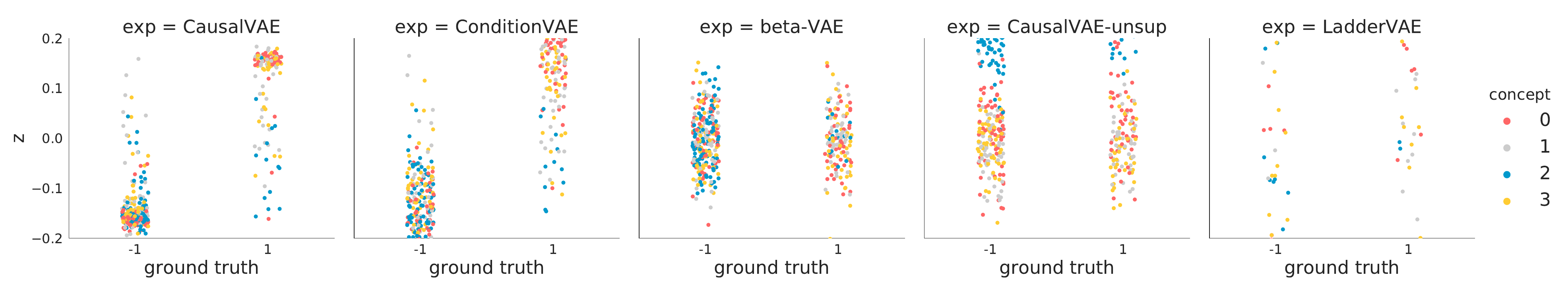}}
\caption{The figure shows the alignment between ground truth $p(\mathbf{z}|\mathbf{u})$ and the learned latent factors $q(\mathbf{z}|\mathbf{x,u})$ on CelebA for 5 methods (CausalVAE, ConditionVAE, $\beta$-VAE, CausalVAE-unsup, LadderVAE from left to right). The ground truth is a  distribution with mean taken from $\{-1,1\}$, and the color of the points indicates different dimensions. The factors learned by CausalVAE show the best alignment among all. The concepts include: 1 \textsc{gender}; 2 \textsc{smile}; 3 \textsc{eyes open}; 4 \textsc{mouth open}.}
\label{align_smile}
\end{center}
\end{figure*}

\subsection{The Learned Graph}
\begin{figure*}[htb]
\centering
\subfigure[Initialize]{
\begin{minipage}{0.45\columnwidth}
\centering
\includegraphics[width=1\textwidth]{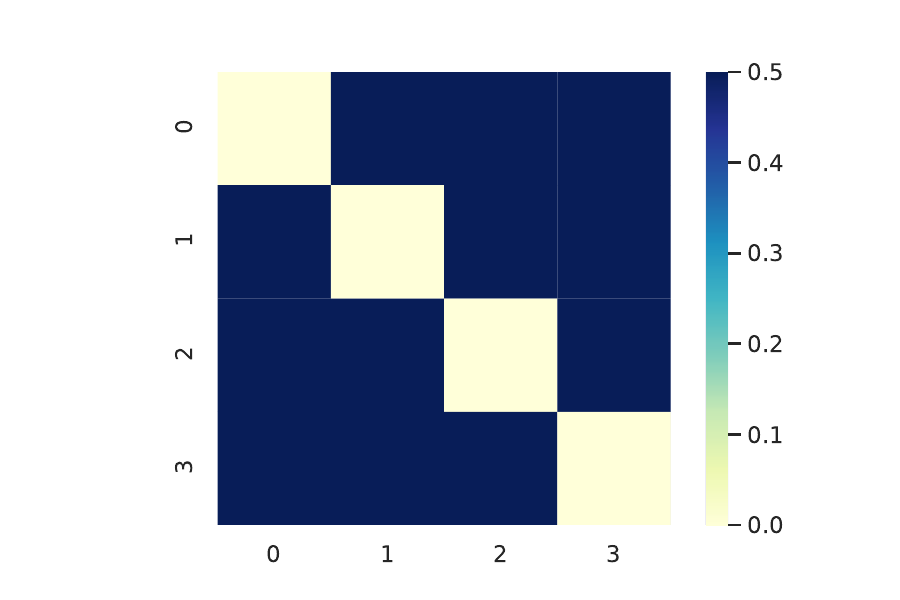}
\end{minipage}%
}
\subfigure[After 1 epoch]{
\begin{minipage}{0.45\columnwidth}
\centering
\includegraphics[width=1\textwidth]{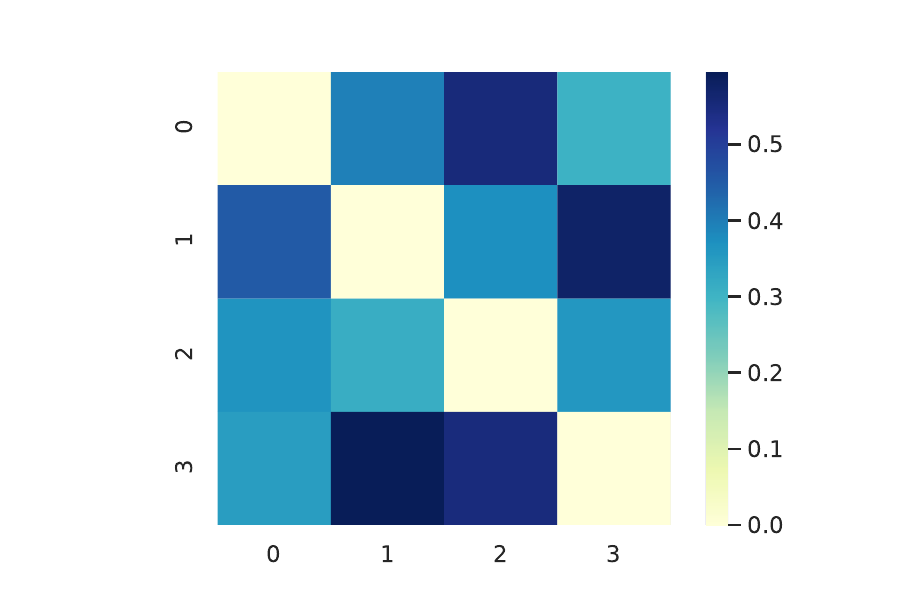}
\end{minipage}%
}
\subfigure[After 5 epochs]{
\begin{minipage}{0.45\columnwidth}
\centering
\includegraphics[width=1\textwidth]{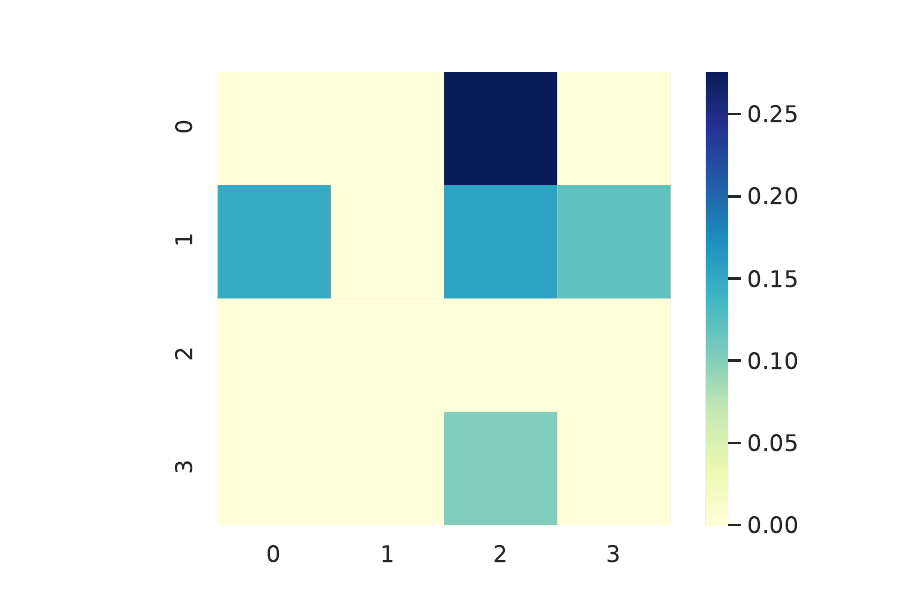}
\end{minipage}
}
\subfigure[The true]{
\begin{minipage}{0.45\columnwidth}
\centering
\includegraphics[width=1\textwidth]{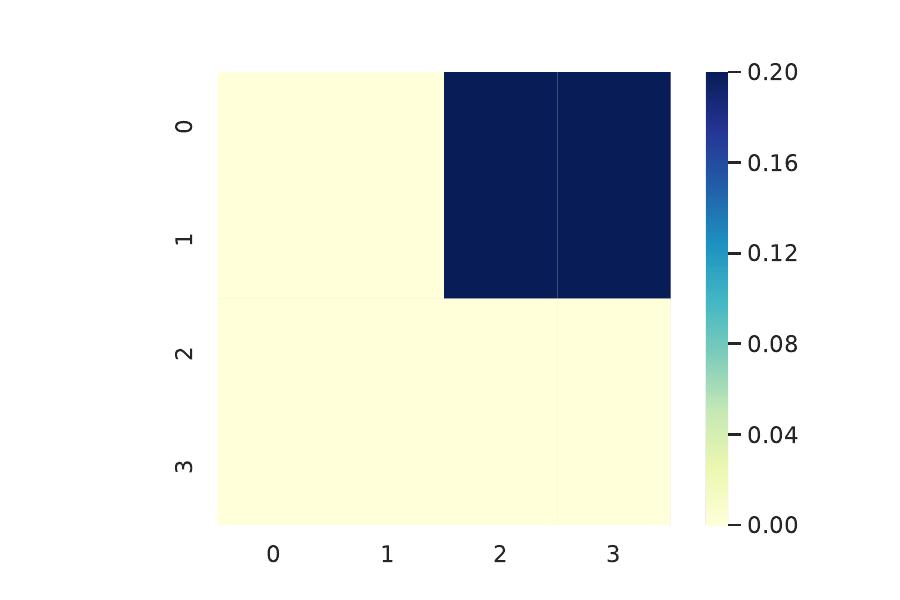}
\end{minipage}%
}
\caption{Learning process of causal graph $\mathbf{A}$ in CelebA (\textsc{Beard}). The concepts include: 1 \textsc{age}; 2 \textsc{gender}; 3 \textsc{bald}; 4 \textsc{beard}.}
\label{beard_graph}
\end{figure*}

\begin{figure*}[h]
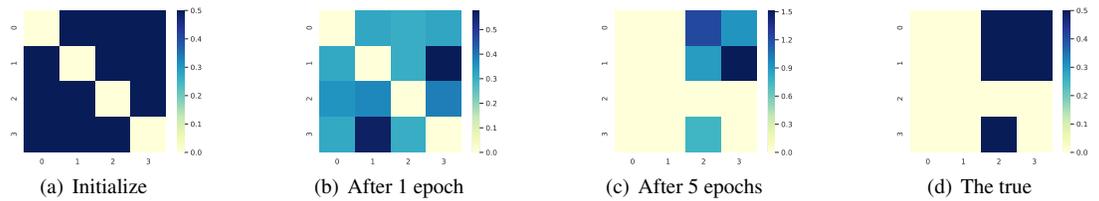

\centering
\subfigure[Initialize]{
\begin{minipage}{0.45\columnwidth}
\centering
\includegraphics[width=1\textwidth]{result_graph/smile/smile_1.pdf}
\end{minipage}%
}
\subfigure[After 1 epoch]{
\begin{minipage}{0.45\columnwidth}
\centering
\includegraphics[width=1\textwidth]{result_graph/smile/smile_2.pdf}
\end{minipage}%
}
\subfigure[After 5 epochs]{
\begin{minipage}{0.45\columnwidth}
\centering
\includegraphics[width=1\textwidth]{result_graph/smile/smile_3.pdf}
\end{minipage}
}
\subfigure[The true]{
\begin{minipage}{0.45\columnwidth}
\centering
\includegraphics[width=1\textwidth]{result_graph/smile/smile_true.pdf}
\end{minipage}%
}
\caption{Learning process of causal graph $\mathbf{A}$ in CelebA (\textsc{Smile}). The concepts include: 1 \textsc{gender}; 2 \textsc{smile}; 3 \textsc{eyes open}; 4 \textsc{mouth open}.}
\label{smile_graph}
\end{figure*}
We demonstrate the learning process of causal graph in this section. Fig. \ref{beard_graph} shows the graph learned process of CelebA (\textsc{Beard}). In this process, we initialize all the entries in $\mathbf{A}$ as 0.5. After 5 epochs, the graph converges. We  observe an almost correct graph in this group of concepts.

\subsection{Intervention Results}
\begin{figure}[H]

\begin{center}
\centerline{\includegraphics[width=0.5\columnwidth]{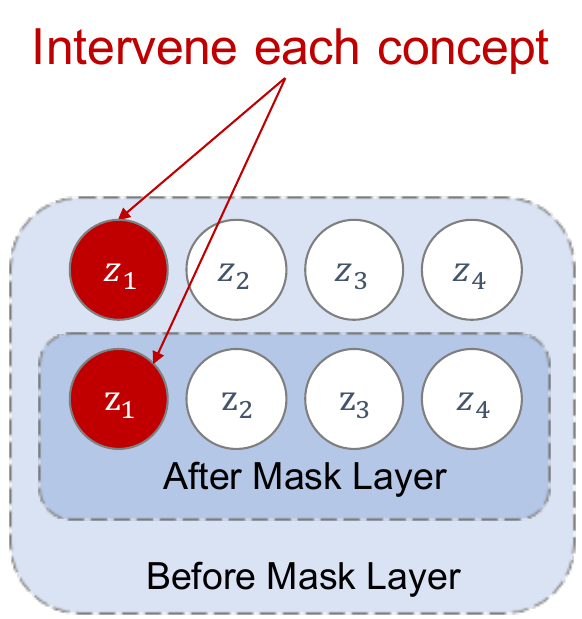}}
\caption{Intervention method}
\label{decodernet}
\end{center}

\end{figure}
The intervention operations are as:
\begin{itemize}
    \item For the learned model, we first put an random observed image $\mathbf{x}$ into the encoder. In this process we could get $\bm{\epsilon}$ and $\mathbf{z}$.
    \item Then for i-th concept, we fix the value of $z_i$ and $g_i({\mathbf{A_i\circ z}})$ as constants. 
    \item Finally, we put the new $\mathbf{z}$ into the decoder and get $\mathbf{x'}$.
\end{itemize}

\begin{figure*}[htb]
\centering
\subfigure[\textsc{age}]{
\begin{minipage}{0.9\columnwidth}
\centering
\includegraphics[width=1\textwidth]{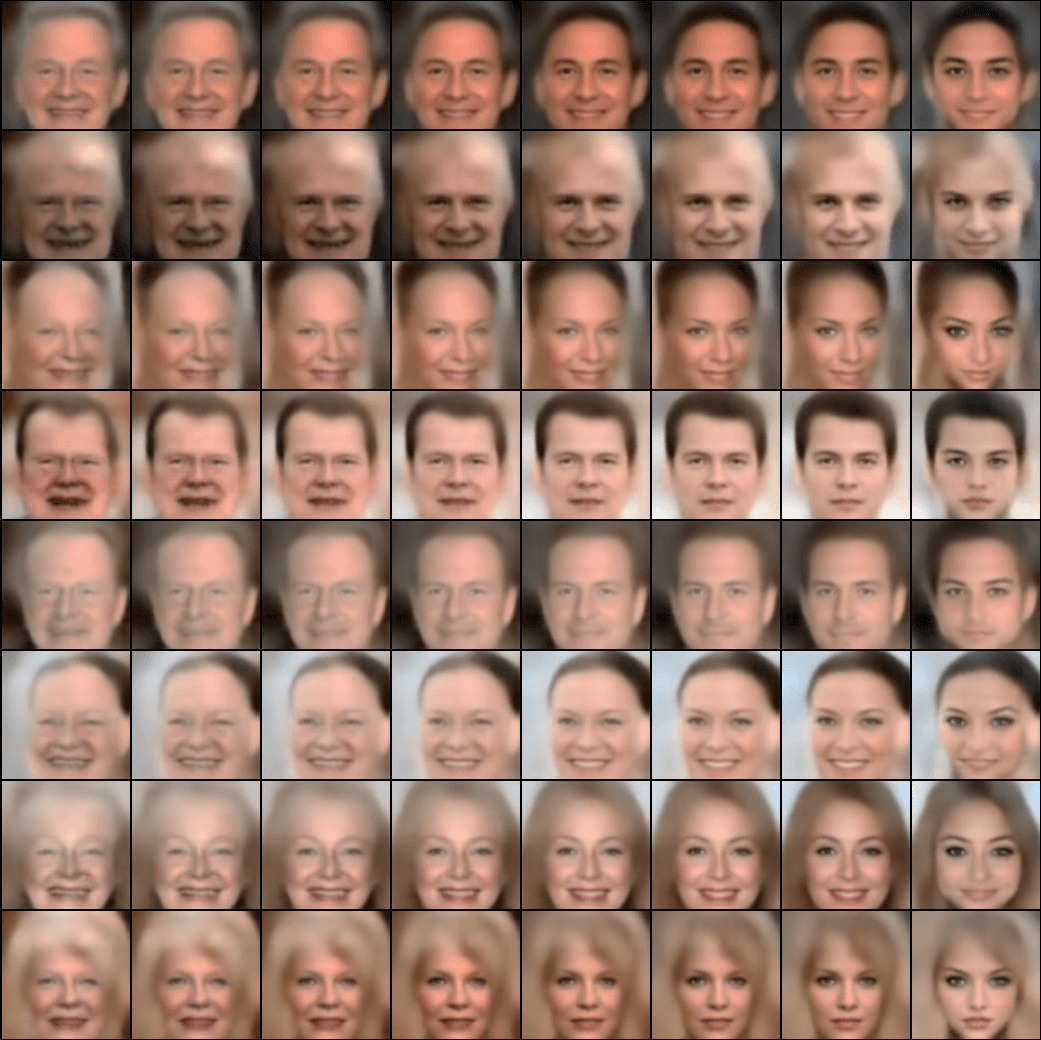}
\end{minipage}%
}
\subfigure[\textsc{gender}]{
\begin{minipage}{0.9\columnwidth}
\centering
\includegraphics[width=1\textwidth]{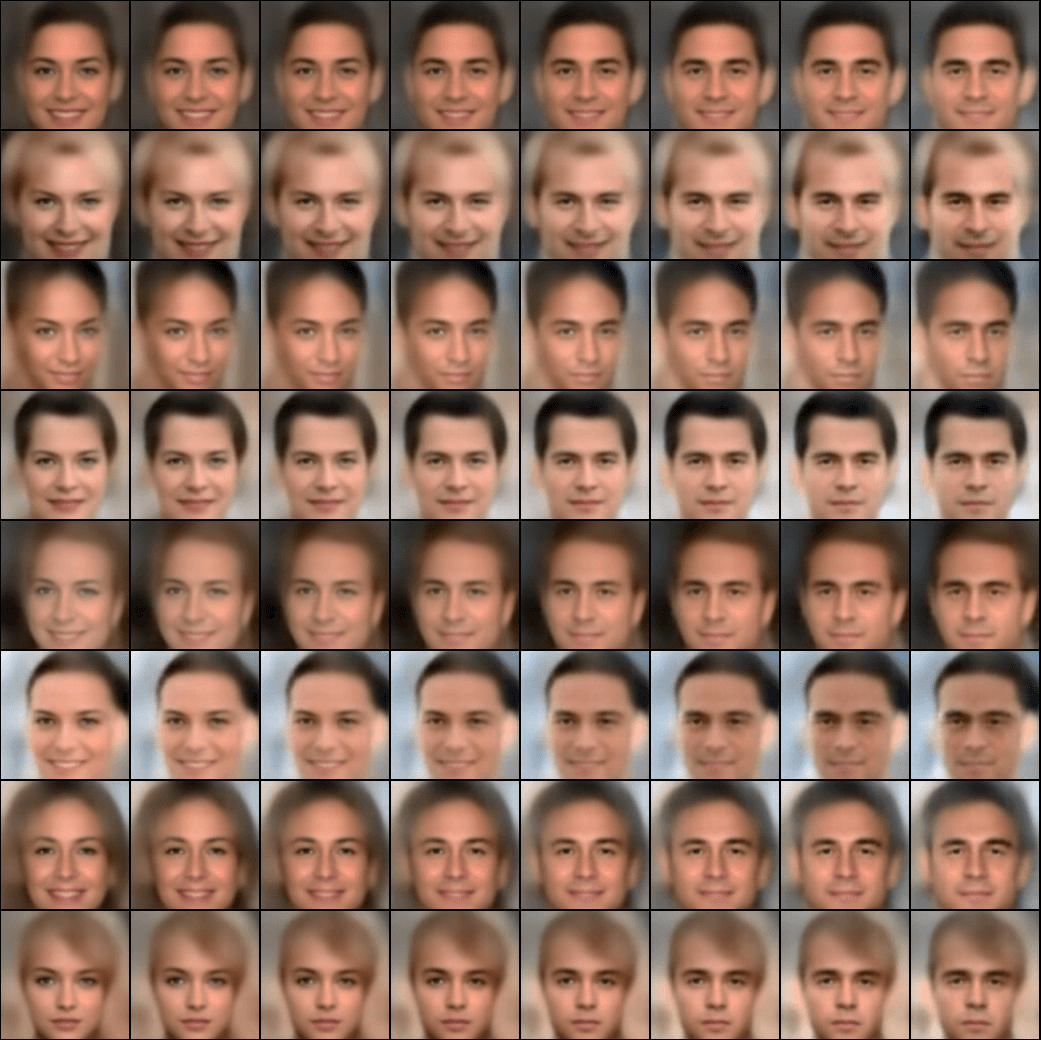}
\end{minipage}%
}
\subfigure[\textsc{bald}]{
\begin{minipage}{0.9\columnwidth}
\centering
\includegraphics[width=1\textwidth]{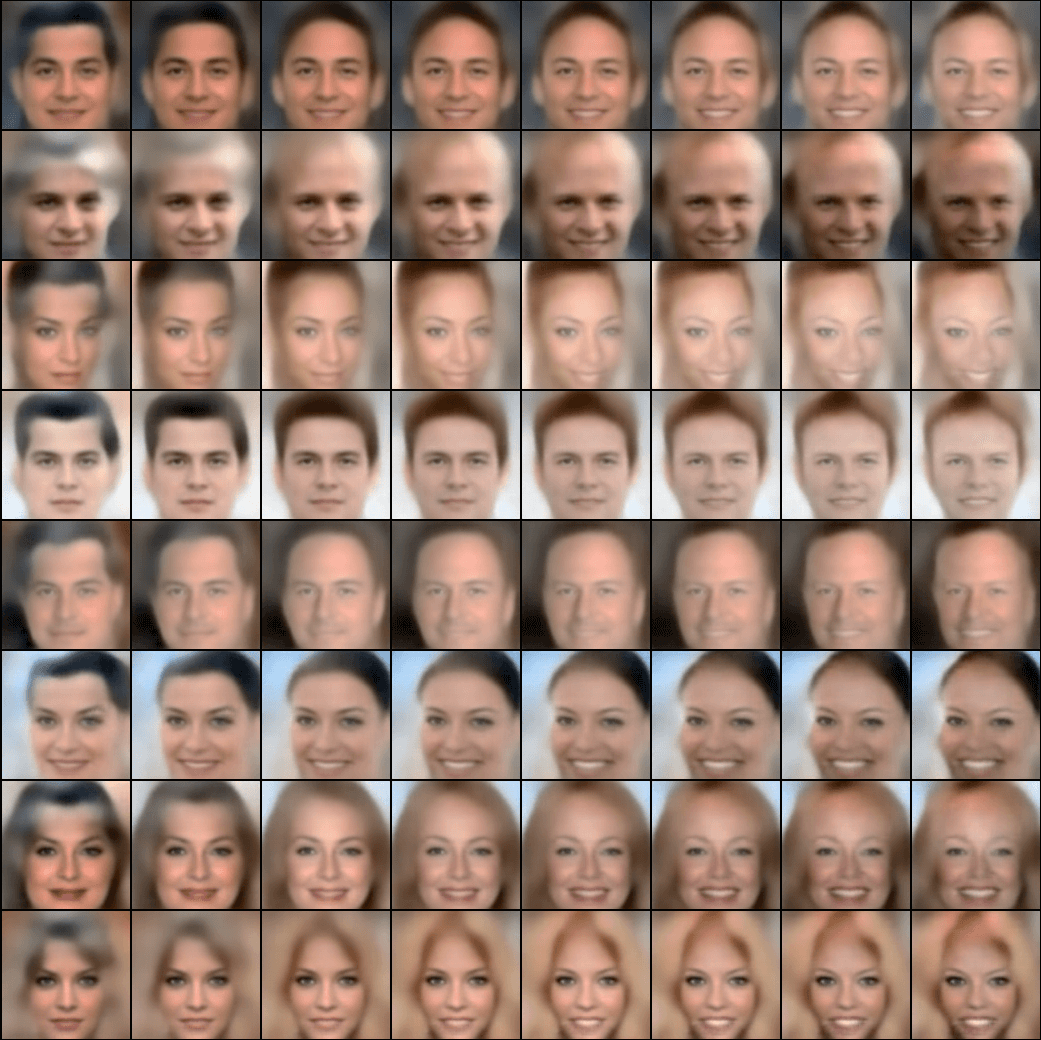}
\end{minipage}%
}
\subfigure[\textsc{beard}]{
\begin{minipage}{0.9\columnwidth}
\centering
\includegraphics[width=1\textwidth]{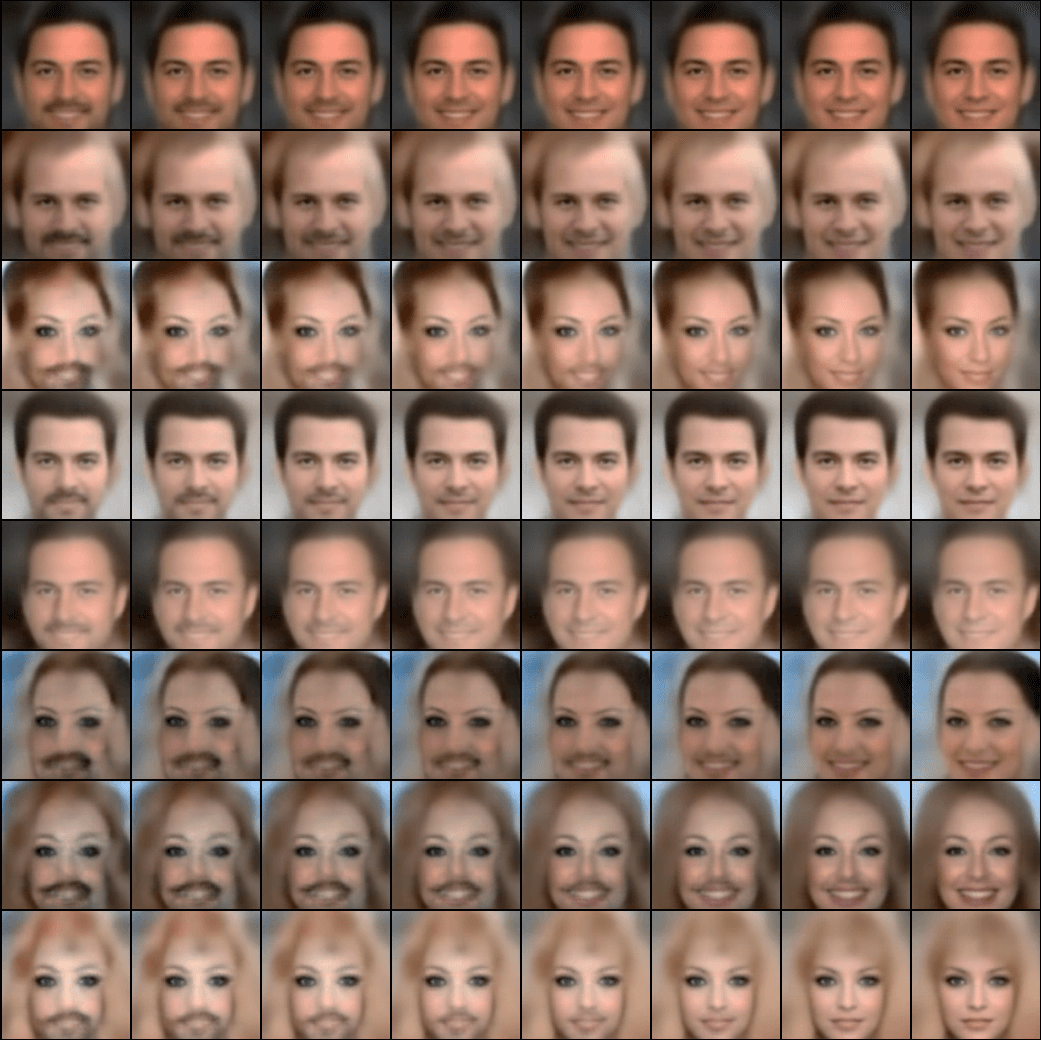}
\end{minipage}%
}
\caption{Results of CausalVAE model on CelebA (\textsc{Beard}). The captions of the subfigures describe the controlled factors. From left to right, the pictures are results obtained by varying the value of the controlled factors.}
\label{face_res_1}
\end{figure*}

\begin{figure*}[htb]
\centering
\subfigure[\textsc{age}]{
\begin{minipage}{0.9\columnwidth}
\centering
\includegraphics[width=1\textwidth]{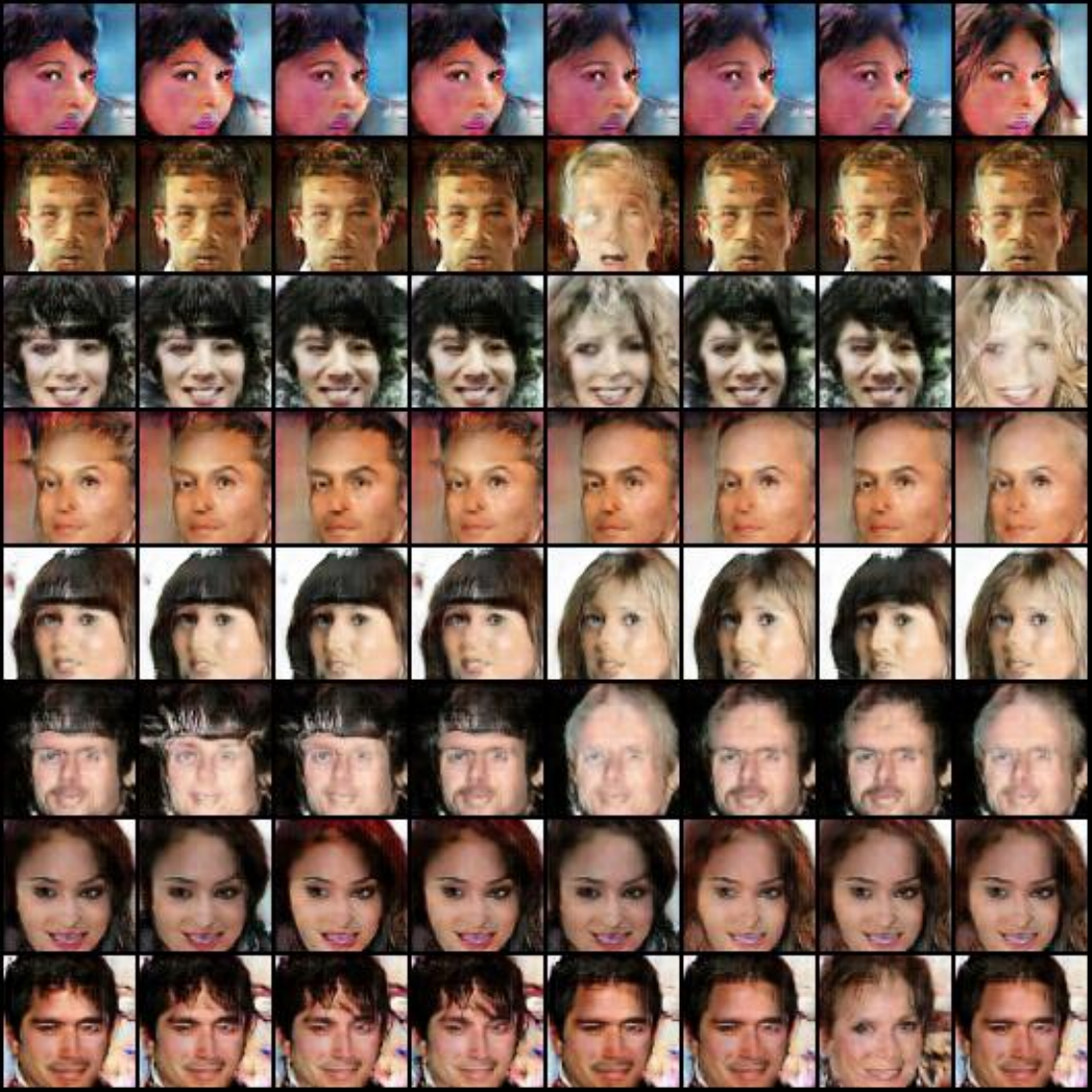}
\end{minipage}%
}
\subfigure[\textsc{gender}]{
\begin{minipage}{0.9\columnwidth}
\centering
\includegraphics[width=1\textwidth]{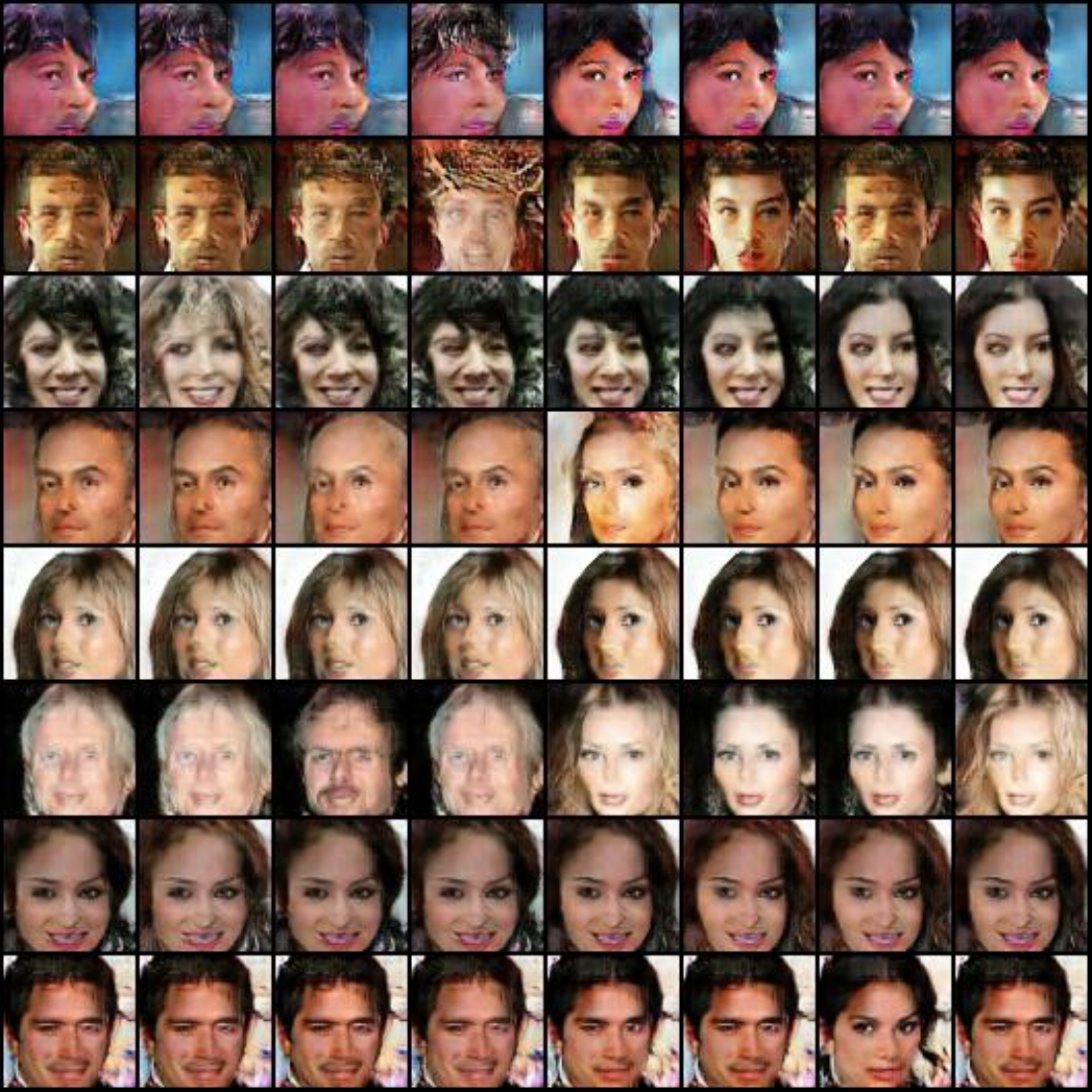}
\end{minipage}%
}
\subfigure[\textsc{bald}]{
\begin{minipage}{0.9\columnwidth}
\centering
\includegraphics[width=1\textwidth]{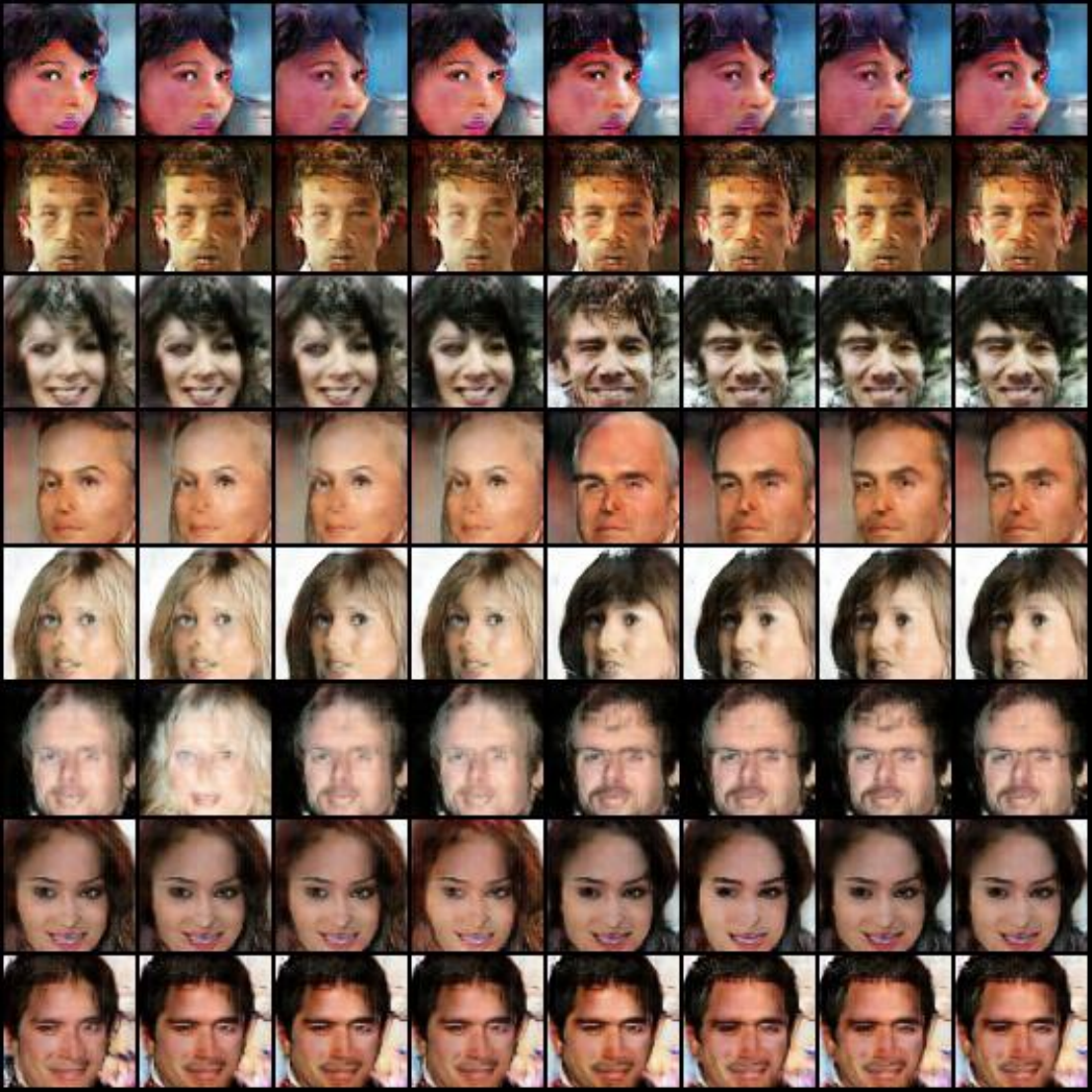}
\end{minipage}%
}
\subfigure[\textsc{beard}]{
\begin{minipage}{0.9\columnwidth}
\centering
\includegraphics[width=1\textwidth]{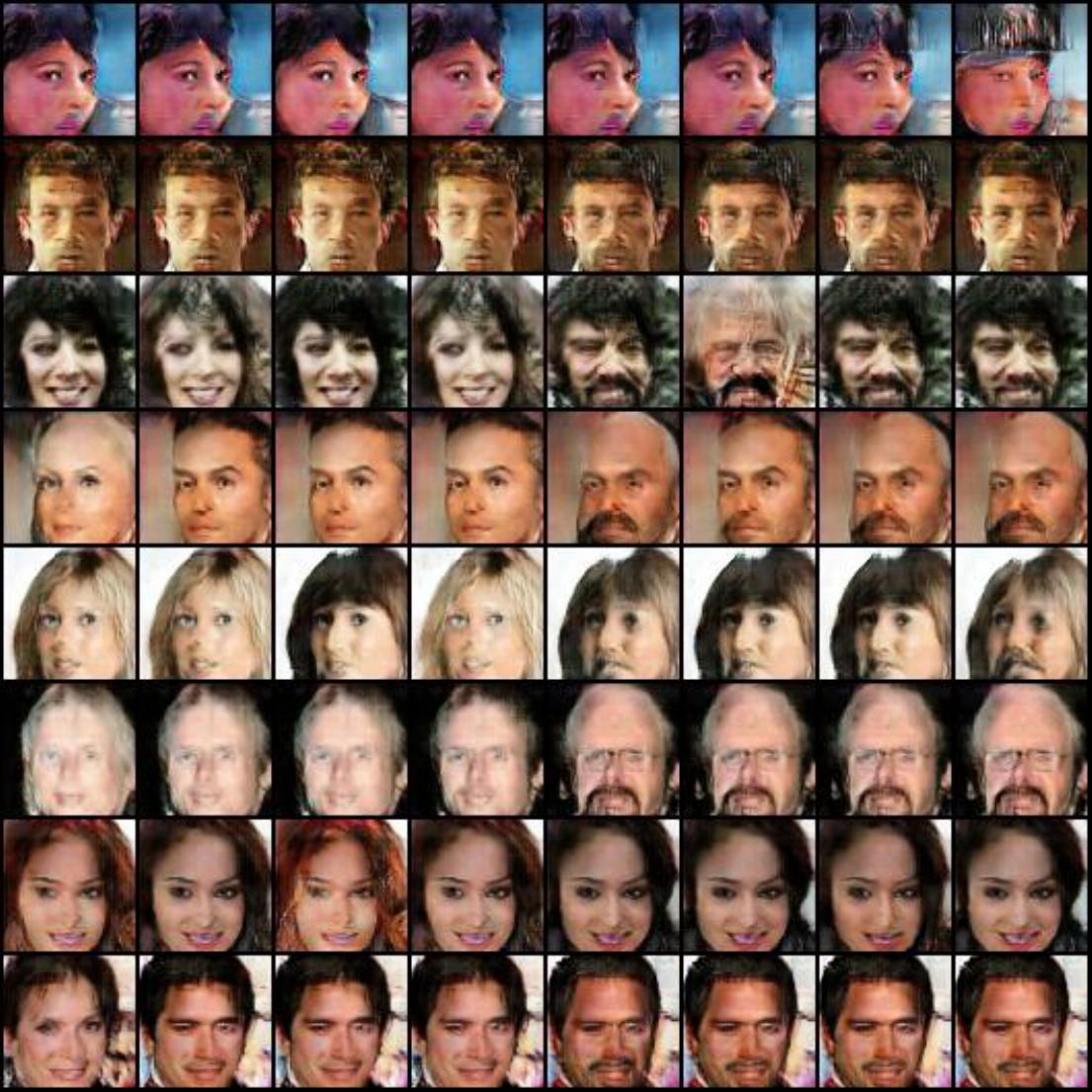}
\end{minipage}%
}
\caption{Results of CausalGAN \cite{causalgan} model on CelebA (\textsc{Beard}). The captions of the subfigures describe the controlled factors. From left to right, the pictures are results obtained by varying the value of the controlled factors.}
\label{ganbeard}
\end{figure*}

Fig. \ref{flow_res} (a) demonstrates the intervention results of CausalVAE on Flow dataset. We see that when we intervene on the cause concept \textsc{ball size}, its child concepts \textsc{water height} and \textsc{water flow} change correspondingly. Similarly, when the cause concept \textsc{hole} is intervened, its child concept \textsc{water flow} also changes. In contrast, intervening on effect concept \textsc{water height} does not influence the causal concept \textsc{ball size}.
Fig. \ref{flow_res}(b) shows the results of ConditionVAE on Flow. We observe that when we intervene on \textsc{ball size}, \textsc{water height} and \textsc{water flow} are affected as expected. However when we intervene on the effect concepts \textsc{water height} and \textsc{water flow}, concept \textsc{ball size} is also influenced, which makes no sense. In general,  the ``do-intervention" of ConditionVAE performs worse than CausalVAE. The results support that by simply using a supervised model, one can not guarantee a causal disentangled representation.

The Fig. \ref{face_res_1} demonstrates the result of CausalVAE on real world banchmark dataset CelebA (\textsc{Beard}), with subfigures (a) (b) (c) (d) showing the intervention experiments on concepts of \textsc{age}, \textsc{gender}, \textsc{bald} and \textsc{beard} respectively. The interventions perform well that when we intervened the cause concept \textsc{gender}, the \textsc{beard} changes correspondingly. Similarly, when the cause concept \textsc{age} in intervened, its child concept \textsc{bald} also changes. In contrast, intervening effect concept \textsc{beard} does not influence the causal concepts \textsc{gender} and other unrelated concepts in Fig. \ref{face_res_1} (d).  Fig. \ref{ganbeard} demonstrates the results of CausalGAN, with subfigures (a) (b) (c) (d) showing the intervention experiments on concepts CelebA (\textsc{Beard}). We observe that when we intervene \textsc{gender}, the \textsc{beard} are changed. But when we intervene \textsc{beard}, concept \textsc{gender} is also changed in third line as shown by Fig. \ref{ganbeard} (d). In general,  the 'do-intervention' of CausalGAN performs worse than CausalVAE.

\begin{figure*}[htb]
\centering
\subfigure[\textsc{gender}]{
\begin{minipage}{0.9\columnwidth}
\centering
\includegraphics[width=1\textwidth]{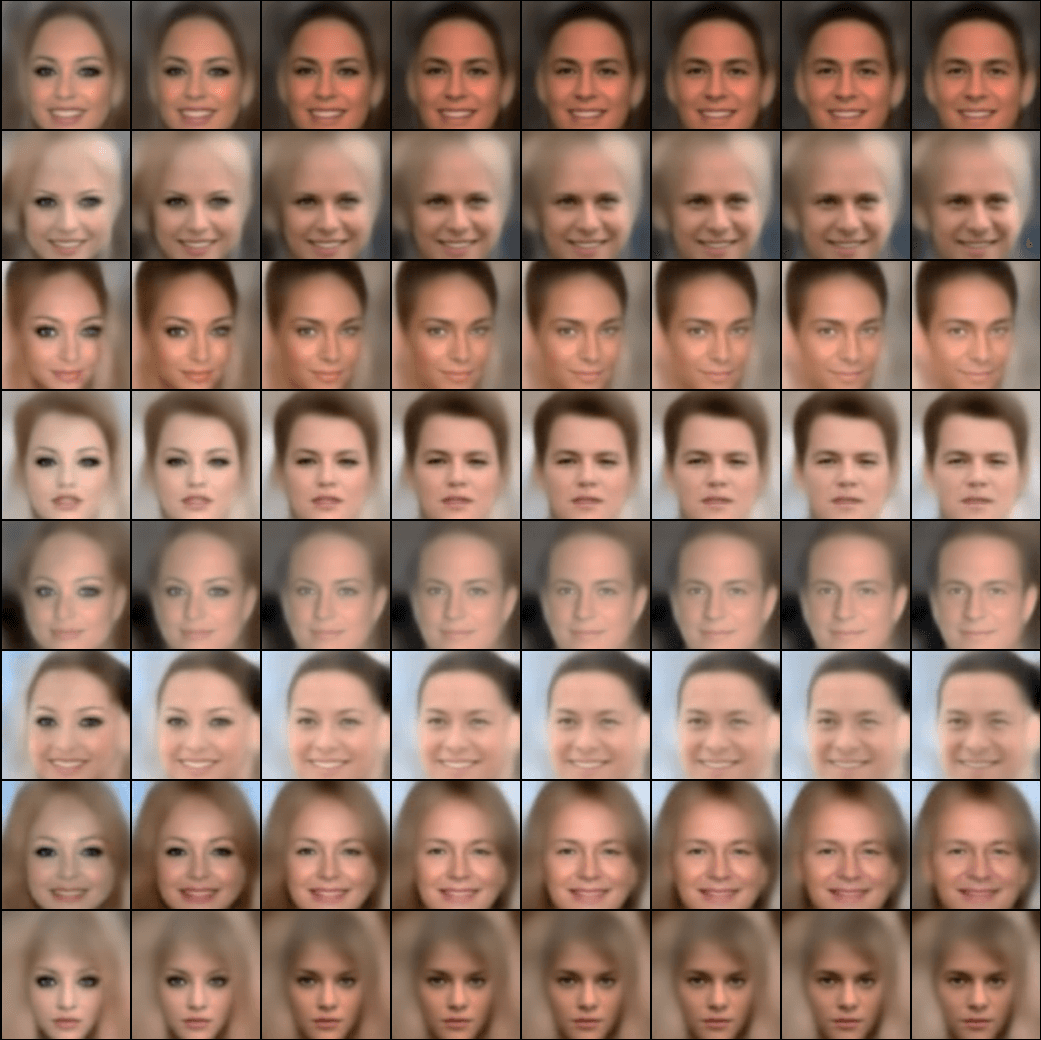}
\end{minipage}%
}
\subfigure[\textsc{smile}]{
\begin{minipage}{0.9\columnwidth}
\centering
\includegraphics[width=1\textwidth]{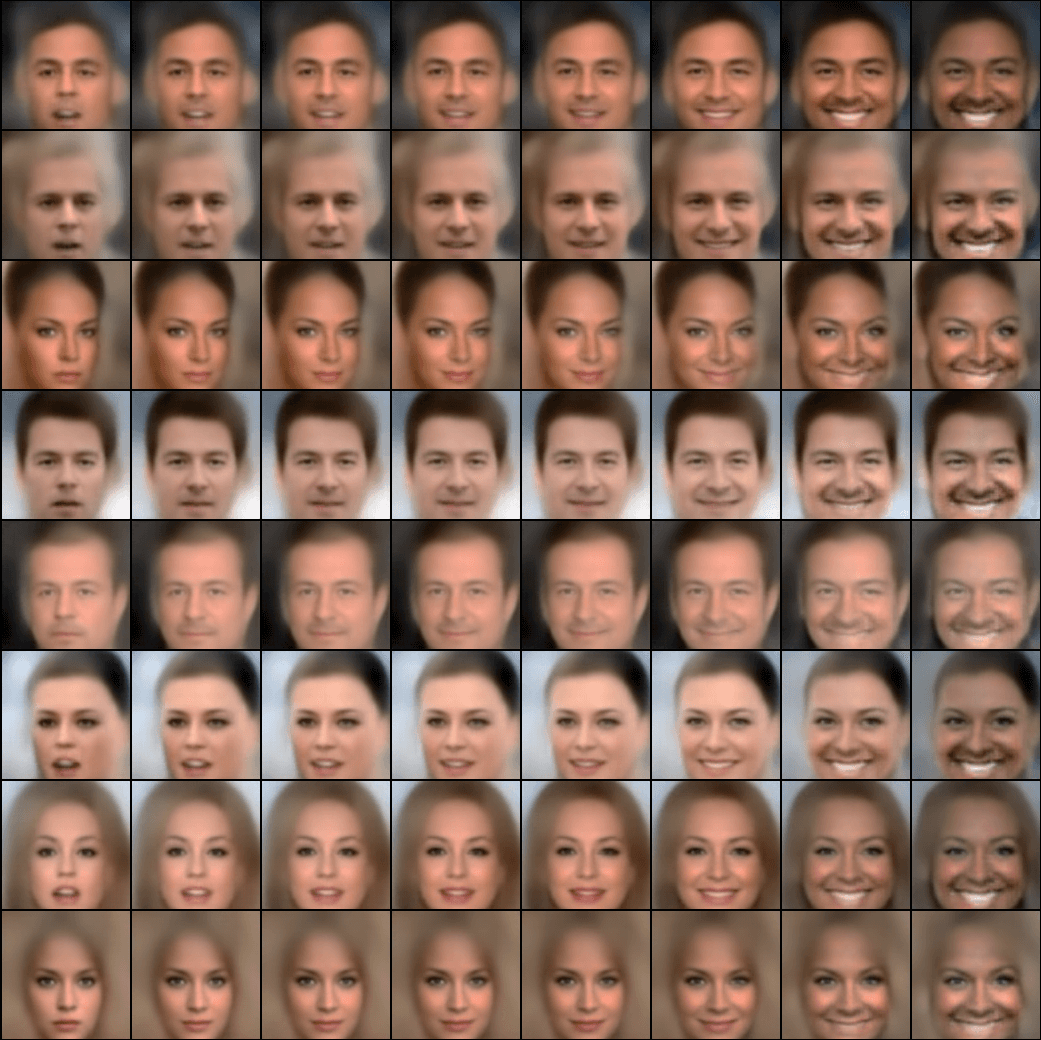}
\end{minipage}%
}
\subfigure[\textsc{eyes open}]{
\begin{minipage}{0.9\columnwidth}
\centering
\includegraphics[width=1\textwidth]{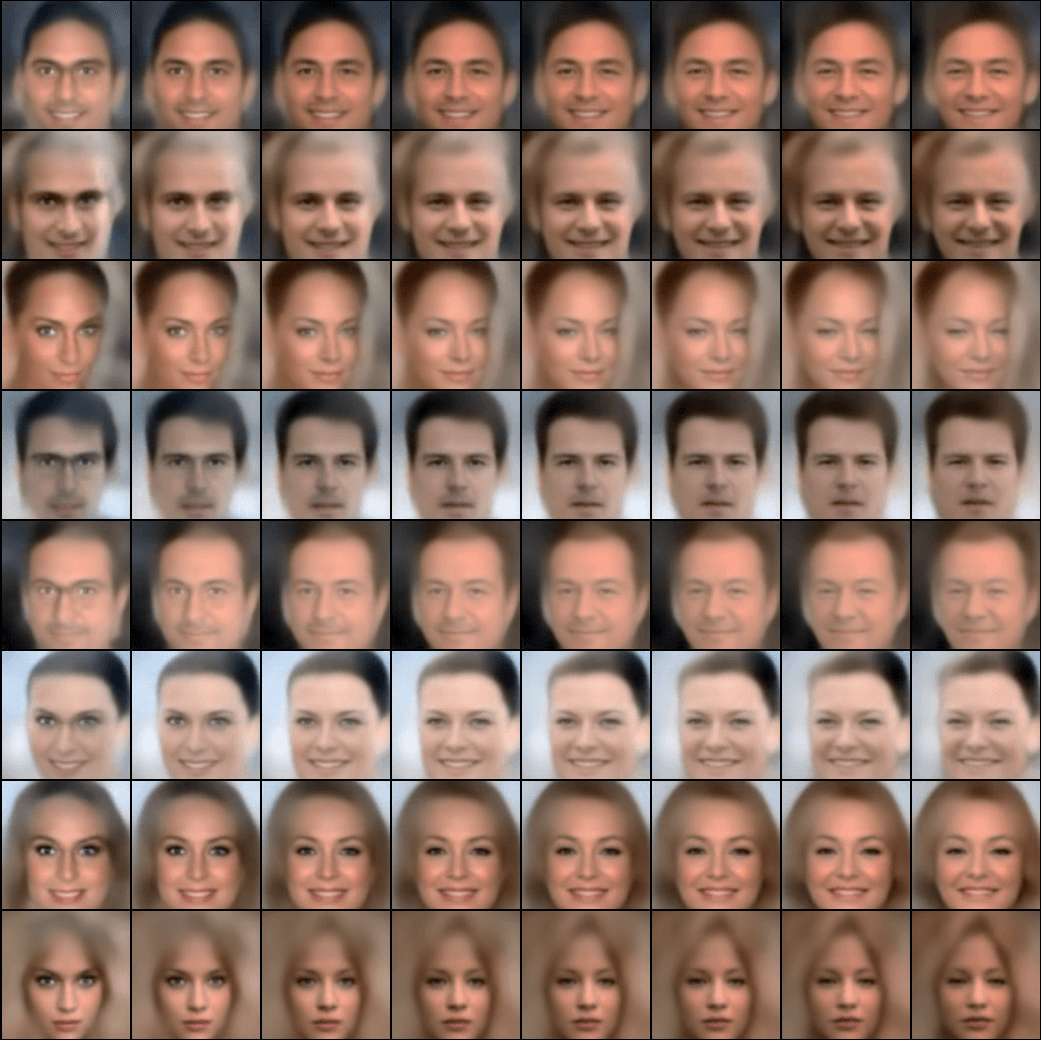}
\end{minipage}%
}
\subfigure[\textsc{mouth open}]{
\begin{minipage}{0.9\columnwidth}
\centering
\includegraphics[width=1\textwidth]{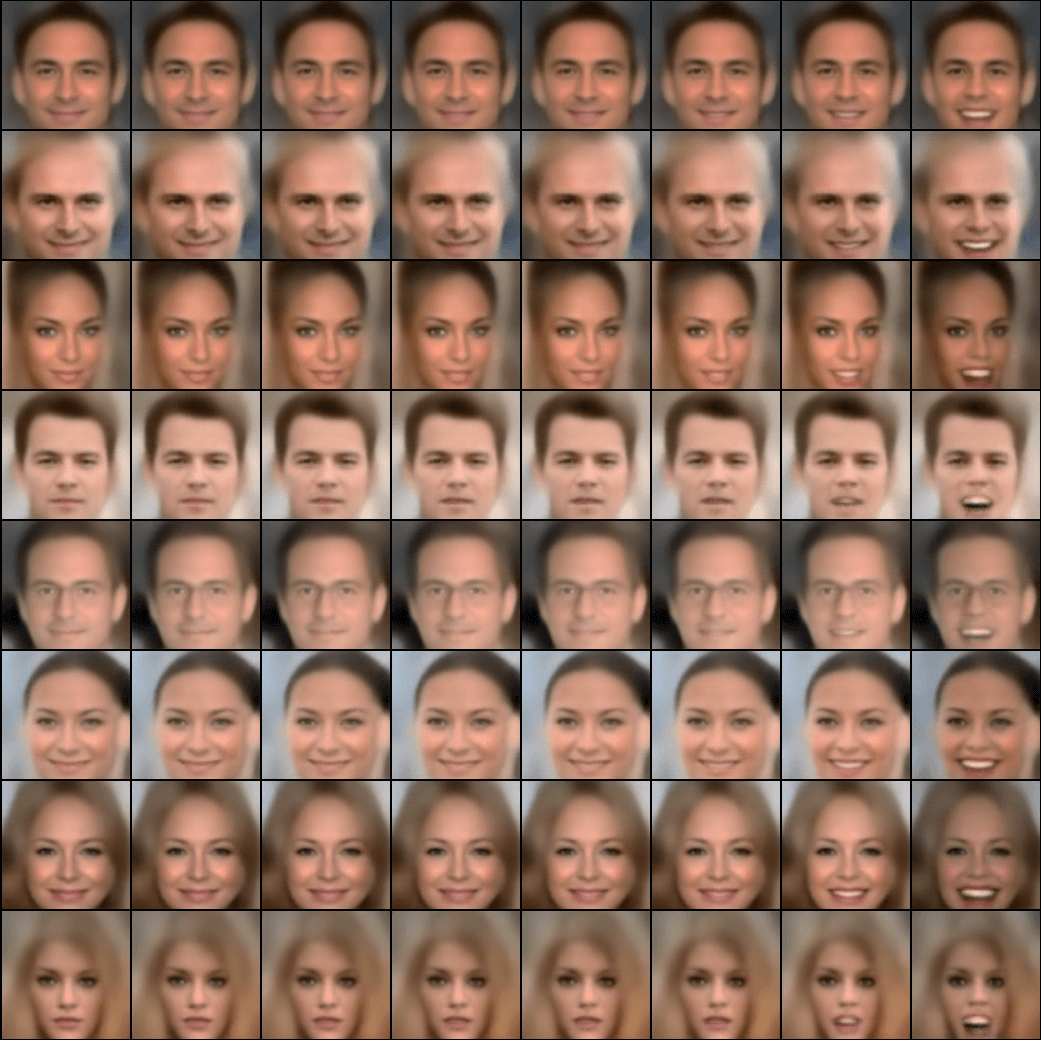}
\end{minipage}%
}
\caption{Results of CausalVAE model on CelebA (\textsc{Smile}). The captions of the subfigures describe the controlled factors. From left to right, the pictures are results obtained by varying the value of the controlled factors.}
\label{face_res_2}
\end{figure*}

\begin{figure*}[htb]
\centering
\subfigure[\textsc{gender}]{
\begin{minipage}{0.9\columnwidth}
\centering
\includegraphics[width=1\textwidth]{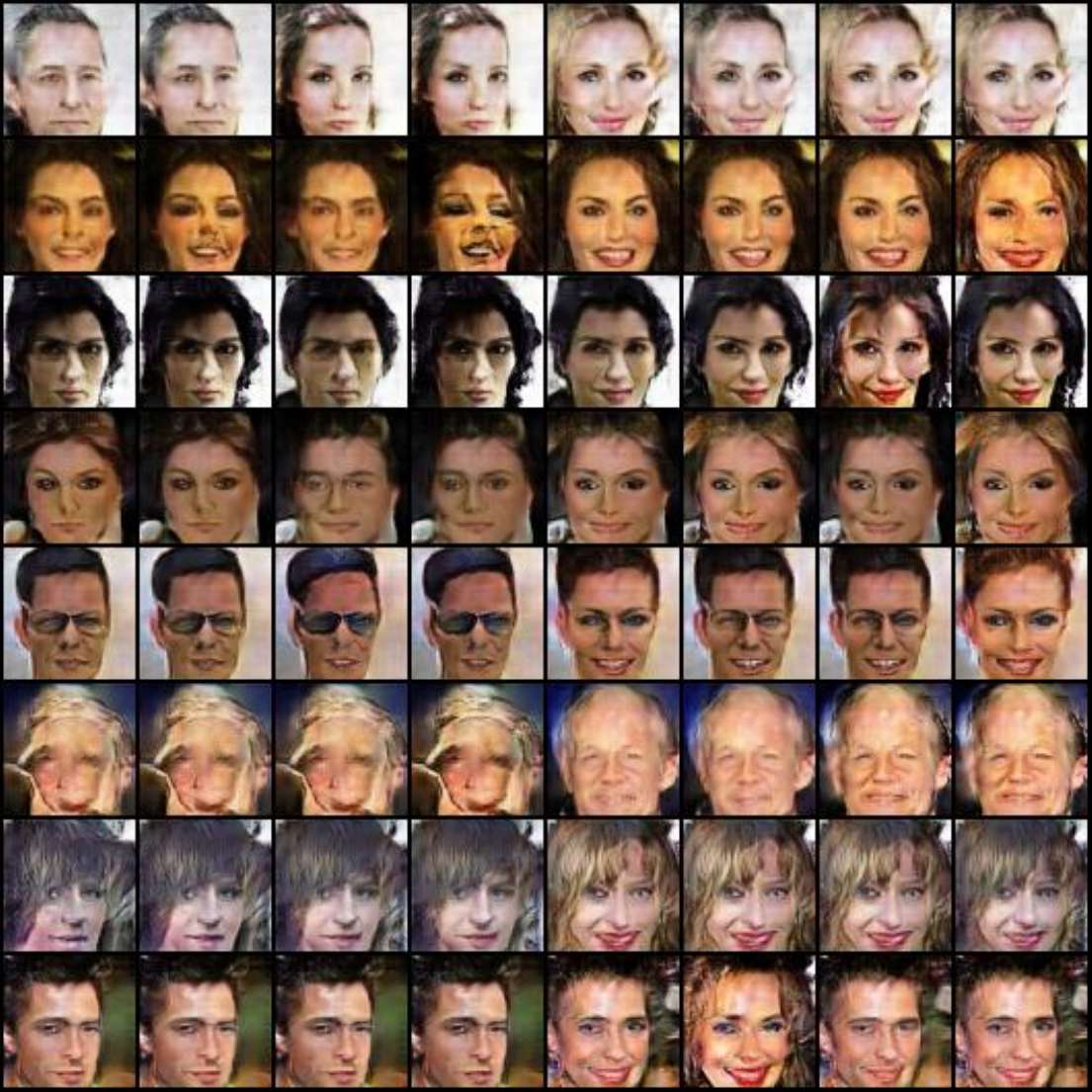}
\end{minipage}%
}
\subfigure[\textsc{smile}]{
\begin{minipage}{0.9\columnwidth}
\centering
\includegraphics[width=1\textwidth]{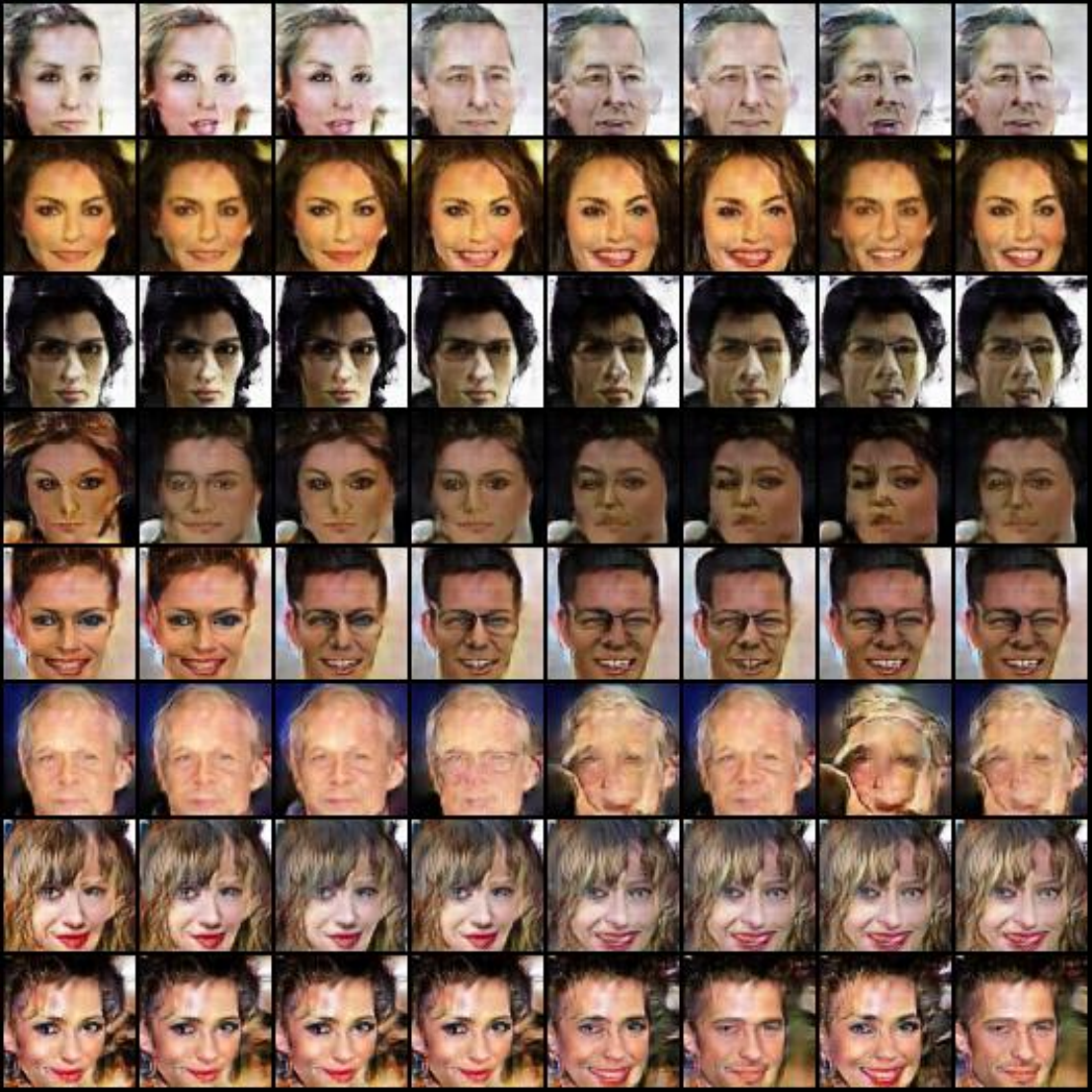}
\end{minipage}%
}
\subfigure[\textsc{eyes open}]{
\begin{minipage}{0.9\columnwidth}
\centering
\includegraphics[width=1\textwidth]{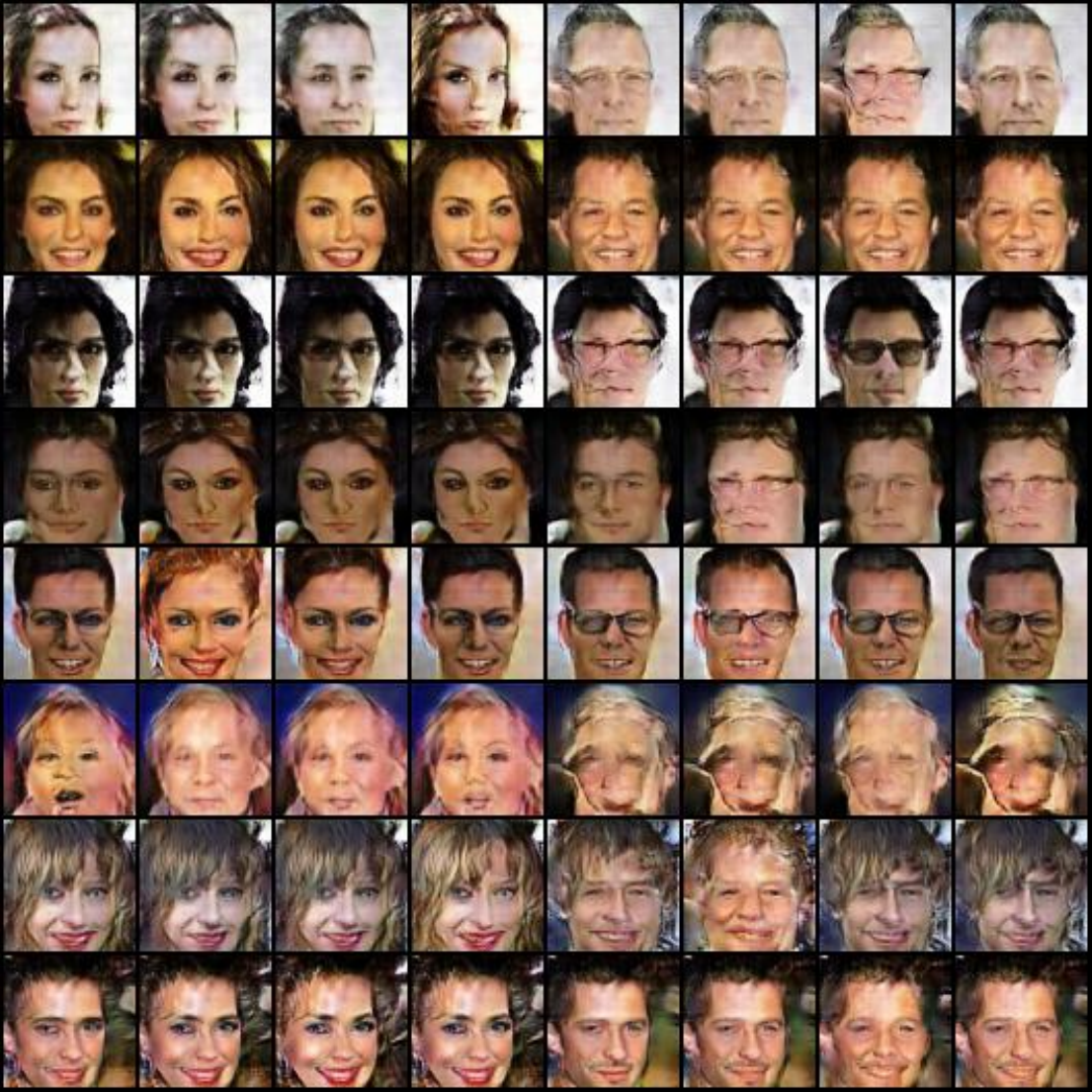}
\end{minipage}%
}
\subfigure[\textsc{mouth open}]{
\begin{minipage}{0.9\columnwidth}
\centering
\includegraphics[width=1\textwidth]{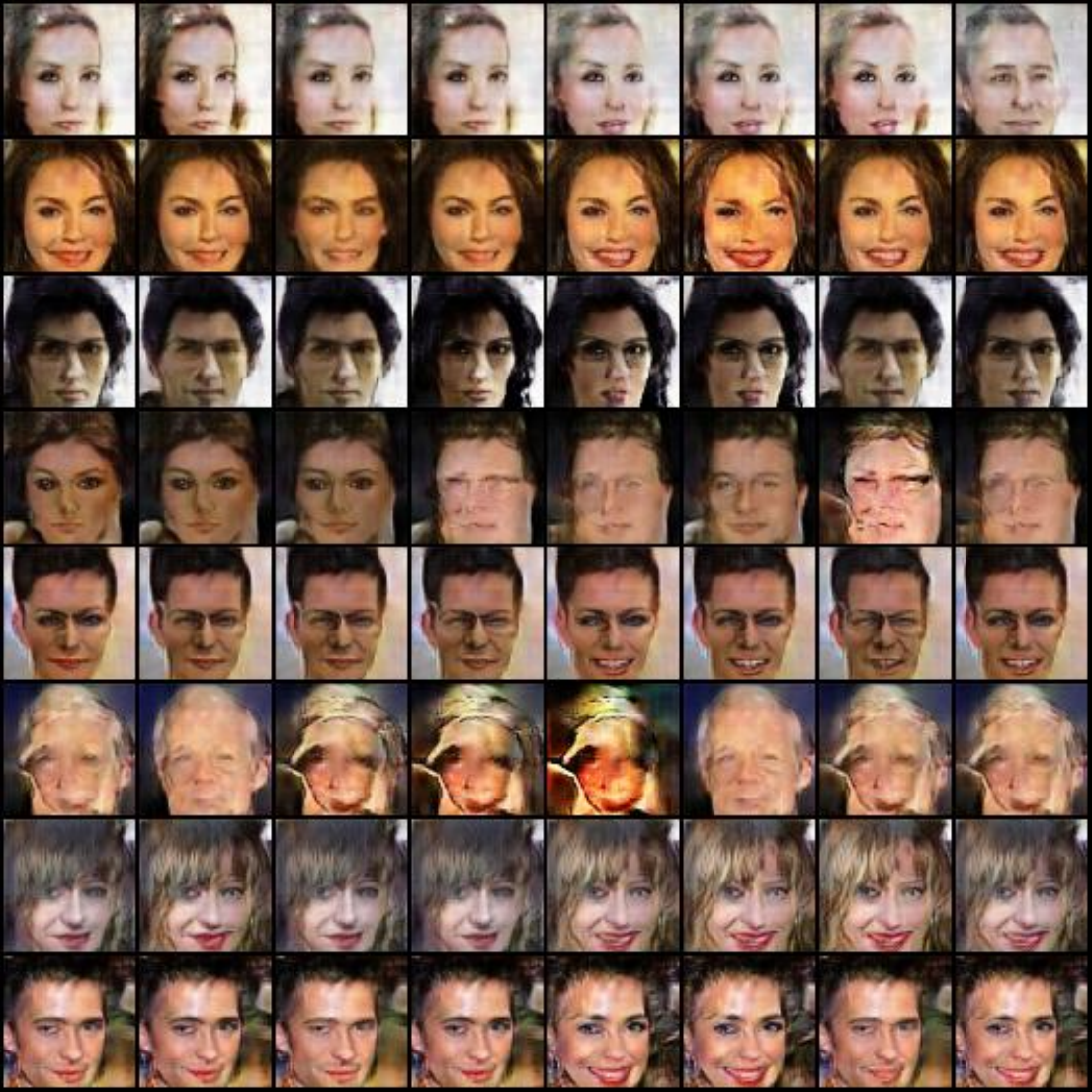}
\end{minipage}%
}
\caption{Results of CausalGAN model on CelebA (\textsc{Smile}). The captions of the subfigures describe the controlled factors. From left to right, the pictures are results obtained by varying the value of the controlled factors.}
\label{gansmile}
\end{figure*}

The Fig. \ref{face_res_2} demonstrates the result of CausalVAE on real world banchmark dataset CelebA (\textsc{Smile}), with subfigures (a) (b) (c) (d) showing the intervention experiments on concepts of \textsc{gender}, \textsc{smile}, \textsc{mouth open} and \textsc{eyes open} respectively. The interventions perform well that when we intervened the cause concept \textsc{gender}, not only the appearance of \textsc{gender} but the eyes changed. When we intervened the cause concept \textsc{smile}, not only the appearance of \textsc{smile} but the \textsc{mouth open}. In contrast, intervening effect concept \textsc{mouth open} does not influence the causal concepts \textsc{smile} in Fig. \ref{face_res_2} (d).  Fig. \ref{gansmile} demonstrates the results of CausalGAN, with subfigures (a) (b) (c) (d) showing the intervention experiments on concepts CelebA (\textsc{Smile}). We find that when we control \textsc{smile}, the mouth is changed, as shown in  the second line of  Fig. \ref{gansmile} (b). But we find sometimes the control of \textsc{smile} influence other unrelated concepts like \textsc{gender} (shown in first line of Fig. \ref{gansmile}  (b)). In this concepts group, CausalGAN also shows relatively unstable intervention experiments compared to that of ours.
 
\clearpage
\bibliography{references}
\bibliographystyle{abbrvnat}